\documentclass[Journal,letterpaper]{ascelike-new}
\usepackage[utf8]{inputenc}
\usepackage[T1]{fontenc}
\usepackage{lmodern}
\usepackage{graphicx}
\usepackage[figurename=Fig.,labelfont=bf,labelsep=period]{caption}
\usepackage{subcaption}
\usepackage{amsmath}
\usepackage{multirow}
\usepackage{newtxtext,newtxmath}
\usepackage[colorlinks=true,citecolor=red,linkcolor=black]{hyperref}
\usepackage{algorithm}
\usepackage[noend]{algpseudocode}
\usepackage{pdflscape}
\usepackage{makecell}
\NameTag{Agapaki, \today}
\begin{document}

\nolinenumbers


\title{GEOMETRIC DIGITAL TWINNING OF INDUSTRIAL FACILITIES: RETRIEVAL OF INDUSTRIAL SHAPES}

\author[1]{Eva Agapaki}
\author[2]{Ioannis Brilakis}

\affil[1]{Assistant Professor, M.E. Rinker, Sr. School of Construction Management, University of Florida, 32608, U.S.A. Email: agapakie@ufl.edu}
\affil[2]{Professor, Department of Engineering, University of Cambridge, CB2 1PZ, U.K.}

\maketitle

\begin{abstract}
This paper devises, implements and benchmarks a novel shape retrieval method that can accurately match individual labelled point clusters (instances) of existing industrial facilities with their respective CAD models. It employs a combination of image and point cloud deep learning networks to classify and match instances to their geometrically similar CAD model. It extends our previous research on geometric digital twin generation from point cloud data, which currently is a tedious, manual process. Experiments with our joint network reveal that it can reliably retrieve CAD models at 85.2\% accuracy. The proposed research is a fundamental framework to enable the geometric Digital Twin (gDT) pipeline and incorporate the real geometric configuration into the Digital Twin.
\end{abstract}

\section{Introduction}
Our world sees a global infrastructure investment gap and global leaders are working towards closing it. According to the Department of Energy in the United States, the cost of power outages is up to \$70 billion annually leading the country to a required minimum investment of \$65 billion to support development of energy infrastructure \cite{WhiteHouse2022}. The China Belt and Road is another initiative to evaluate long-term strategies that improve infrastructure in multiple continents such as Africa, Europe, Australasia, and Latin America \cite{OECD2018}. The strategy is implemented due to the urgent need to alleviate the excessive amounts of global infrastructure falling short by \$0.35-0.37 trillion per year \cite{GIHub2017,McKinsey2016}. Infrastructure investments are even more significant for fast growing industries, such as the manufacturing industry (with 35\% productivity growth), which creates an outsize economic impact in the United States \cite{McKinsey2021}. Therefore, the high value of manufacturing assets to our economies necessitates the need to properly maintain the industrial sector. 

The United States spends \$74.5 billion per year due to poor maintenance of industrial facilities, and existing documentation is not enough to prevent deterioration of the aging industrial building stock \cite{NIST2020} and reduction of unplanned machinery down-times. The United States Pipeline and Hazardous materials Safety Administration has reported more than 10,000 failures in oil and gas pipelines in the country since 2002, which resulted in \$6 billion financial losses \cite{DOT2016}.  It is essential to highlight that U.S oil plants are running at 82\% of their capacity, reflecting a 10\% decrease in their average production \cite{Reuters2021}. The extended problem results due to cutbacks of maintenance which exacerbates production improvements in the future. In addition to that, there are time constraints on how long equipment and industrial elements can be shut down without affecting production. Production boundaries cannot be crossed without paying additional expenses. These time-critical needs necessitate the use of technological innovations that can alleviate these problems.

Technological innovations have the potential to: (a) capture the detailed geometry of the physical infrastructure and generate the asset’s Digital Twin (DT), (b) enrich the geometric DT with real-time sensor data, (c) update, maintain and communicate with the DT and (d) leverage the DT to monitor the asset’s performance and improve decision-making by planning interventions well before the time of need. Without DTs, the incurred information loss throughout an asset's lifecycle would be immense. The connection to the physical twin is the differentiating factor from any other digital model or replica \cite{IET2019DigitalEnvironment}. The greatest value of DTs in the industrial sector is facilitating preventive maintenance strategies, which entails the accurate identification of industrial/mechanical parts and timely maintenance decisions, and is projected to incur 15-25\% savings to the global infrastructure market by 2025 \cite{Barbosa2017ReinventingProductivity,Gerbert2016DigitalConstruction}, reaching a global market value of \$48.2 billion by 2026 \cite{dtmarket}.

\begin{figure}[!ht]
\centering
\includegraphics[width=\textwidth]{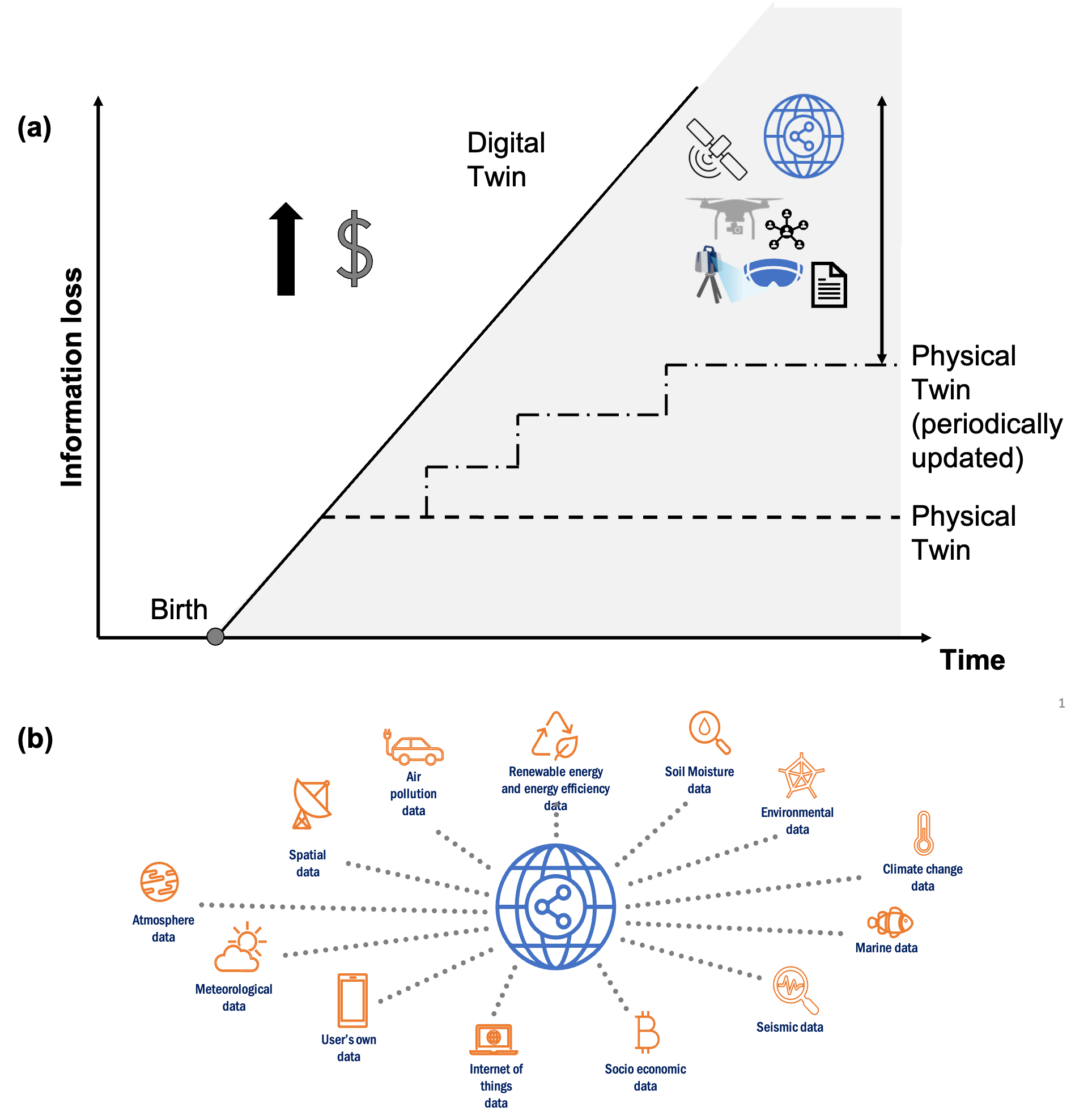}
\caption{(a) Schematic of the information loss in relation to an asset's life (diagram is not to scale) and (b) sample of data sources integrated into the Digital Twin.}
\label{fig:infoloss}
\end{figure}

Digital Twinning is expected to have a significant impact in the Architecture, Engineering and Construction (AEC) industry \cite{sacks2020construction}. DTs integrate ultra-high fidelity simulations of infrastructure assets with their maintenance history and sensor data to mirror the physical assets, thus bringing immense value to all stages of their life-cycle management. This value is currently counterbalanced by the cost and effort currently needed to generate DTs especially for existing and old assets. One of the main cost drivers is processing their complex geometry to a human understandable model. A simplified diagram of the value of DTs starting from the birth of the infrastructure (design phase) throughout its life (demolition phase) is presented in Figure \ref{fig:infoloss}. 

The generation of a geometric Digital Twin (gDT) is the core step of the digital twinning process. Other data can then be linked on top of the gDT (sample data sources shown in Figure \ref{fig:infoloss}(b)), but that process is dependent on accurately capturing the 3D geometry. The current research on gDT generation in either the construction or the operational phases consists of four main steps (1) raw data collection (image and/or point cloud data), (2) data preparation, (3) geometric modeling and (4) semantic enrichment of the model with additional information, such as topological relationships and material specifications. Previous research has identified that 90\% of the gDT generation cost is on the geometric modeling step \cite{Fumarola2011GeneratingApproaches,Hullo2015Multi-SensorArchitectures}. Therefore, the gDT process needs to be automated both from a cost analysis perspective and integration with other industrial systems and existing documentation. In our previous work, we conducted a comprehensive technical assessment and viable evaluation of existing state-of-the-art software tools available \cite{Agapaki2018PrioritizingFacilities} and developed methodologies to automatically partition the input point cloud into clusters \cite{agapaki2020cloi} and individual objects \cite{agapaki2020instance}. In the following paragraphs, we summarize the gDT state-of-the-art (SOTA) practice.

\subsection{State-of-practice}

In our previous work \cite{Agapaki2018PrioritizingFacilities}, we prioritized the most important to automatically model industrial object types based on their frequency of occurrence and labor to model. These are cylinders (straight pipes, electrical conduit and circular hollow sections), valves, elbows, I-beams, angles, channels and flanges. Out of those, cylindrical objects are the most time intensive to model, requiring 80\% of the total modeling time in EdgeWise \cite{ClearEdge2019PlantCapabilities} and represent 45.5\% of the total number of industrial objects on average. 

There has been a bevy of commercial tools that attempt to semi-automatically model the built environment. Recently, PointFuse \cite{PointFuse} powered by JetStream converts point clouds to meshes and then fits 3D standardized shapes at Level Of Detail (LOD) 200, which requires significant manual correction \cite{PointFuse}. Other companies such as VEERUM \cite{Veerum}, Cintoo \cite{Cintoo}, WillowTwin \cite{Willow} and Re-Scan \cite{Rescan} manually tag equipment and machinery to generate asset DTs. The most widely used software modeling packages by Autodesk, Clearedge3D, and AVEVA have tools that allow the 3D modeling of mechanical and piping equipment, however they require highly qualified labor to achieve accurate results and significant human labor. Our previous work used EdgeWise as a comparison, since it has been one of the few commercially available tools to automatically model cylindrical objects from laser scanned point clouds of industrial facilities.  

EdgeWise was selected compared to other SOTA software, because it is the only commercially available tool that attempts to automatically extract cylinders from the point cloud of an industrial plant without significant user assistance. Although EdgeWise has significantly accelerated the 3D modeling of industrial facilities, our previous research summarized some limitations related to fitting standardized geometric shapes (such as cylinders, I-beams) and significant manual modeling time after the automated extraction of geometric shapes \cite{Agapaki2018PrioritizingFacilities}. In particular, cylinder extraction was achieved with 75\% recall and 62\% precision on average.

Therefore, the SOTA practice needs to be improved in the following areas: (a) cylinder segmentation, (b) further classification of cylinders into their sub-categories (i.e., conduits, pipes, circular hollow sections) and (c) automated segmentation and fitting of piping and structural elements (i.e., I-beams, channels, flanges, valves and angles).

\begin{figure}[!ht]
\centering
\includegraphics[width=\textwidth]{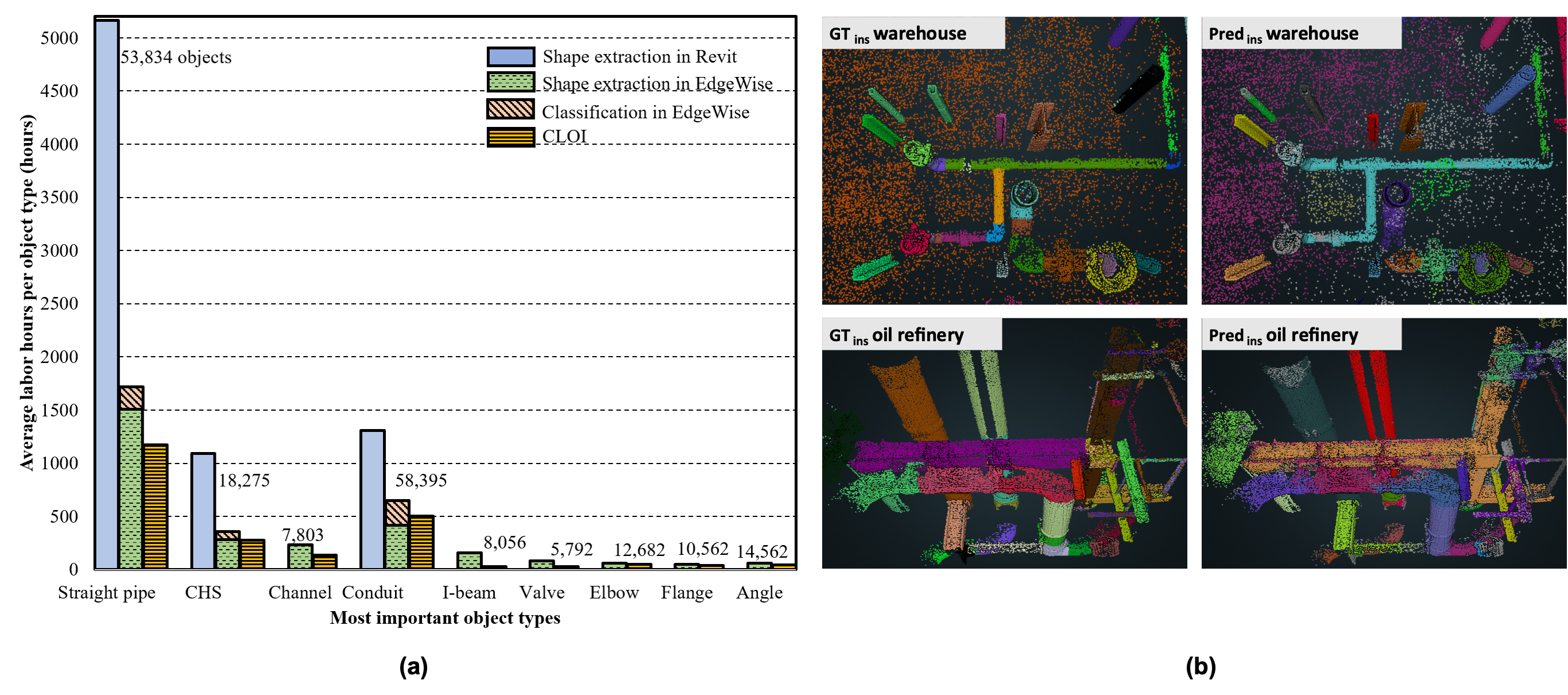}
\caption{(a) Average modeling labor hours per CLOI class for the most important classes of a sample facility with shown number of objects and (b) examples of ground truth and predicted instances of piping element and I-beams.}
\label{fig:cloimodeling}
\end{figure}

Addressing the above-mentioned limitations will improve the tedious modeling practice. We developed a geometric twinning framework, benchmarked it with SOTA practice saving 30\% of modeling hours on average (Figure \ref{fig:cloimodeling}) to minimize manual effort by segmenting point cloud data. This work focuses on fitting geometric shapes and particularly shape retrieval. The sections of this paper are organized as follows. We first present the SOTA research methods that address the above-mentioned limitations. We then define our proposed framework, followed by experiments and results. The last section concludes with discussion of the results and recommendations for future work.

\section{BACKGROUND}
\label{background}

There are multiple gDT generation strategies in the literature. In our previous work, we clustered them into {\bf S1} and {\bf S2} strategies. The {\bf S1} strategy includes (a) primitive industrial shape extraction and (b) fitting. The {\bf S2} strategy comprises of (a) class segmentation, (b) instance segmentation and (c) fitting. Class segmentation entails the process of segregating the 3D point cloud data into semantically meaningful point clusters with similar geometric features. These subsets are called ``classes``. Some examples of industrial object ``classes`` are cylinder, elbow, I-beam, valve, flange, angle and channel point clusters. Instance segmentation is the process of assigning a distinct label per point related to the object that it belongs to. 3D object fitting refers to the process of allocating instance point clusters to geometrically distinct representations. This paper expands upon the {\bf S2} gDT strategy as identified in our previous research \cite{agapaki2020cloi} and particularly the fitting step. 

The primary reasons for selecting this strategy are to avoid data loss due to conversions to other geometric representations (e.g., bounding boxes) and avoid generating standardized geometric representations, which do not take into account the geometric, per-point features. Therefore, the SOTA literature review focuses on a brief review of class and instance segmentation methods and a detailed overview of fitting methods.

\subsection{3D Class segmentation}

\begin{figure}[!ht]
\centering
\includegraphics[width=0.68\textwidth]{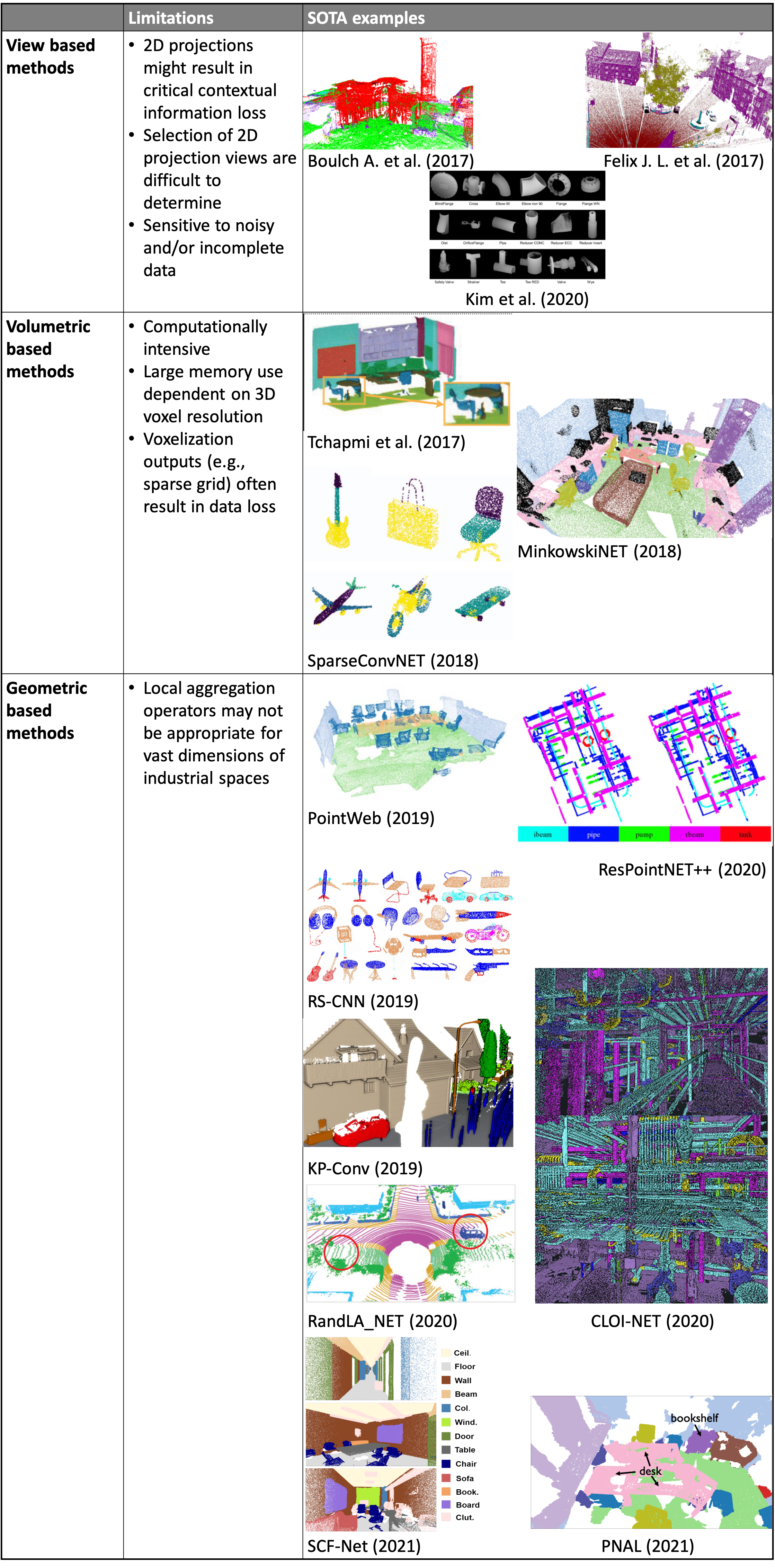}
\caption{SOTA literature review of deep learning based class segmentation methods.}
\label{fig:semseglit}
\end{figure}

Class segmentation of industrial shapes has been widely explored. Our previous research grouped the class segmentation methods into three categories: (a) attribute based, (b) machine learning and (c) deep learning methods. A comprehensive review of those methods can be accessed at \cite{Agapaki2020ChapterGeneration}. The following paragraphs only focus on the SOTA 3D deep learning methods. These methods are grouped into three categories \cite{agapaki2020cloi}. These are view-based, volumetric and geometric deep learning methods. The limitations of each category are summarized in Figure \ref{fig:semseglit}.

View-based methods project the point cloud into multi-view images and then process them by applying Convolutional Neural Networks (CNNs). Then, class segmentation labels are predicted on each 2D projected view and the features from each view are aggregated to obtain class labels per point \cite{boulch2017unstructured,lawin2017deep}. \cite{kim2020deep} used a multi-view CNN (MVCNN) to retrieve catalogs of CAD objects from laser-scanned piping objects (pipes, valves and elbows). However, their method assumes that the piping objects are manually segmented and it does not apply data augmentation to account for the sensitivity of the deep learning-based retrieval. 

Unlike the view-based approaches, volumetric-based methods process voxels or octrees converted from the point cloud. \cite{tchapmi2017segcloud} developed SegCloud, which used a 3D Fully Convolutional CNN (3D-FCNN) to predict coarse voxel scores, mapping them to the raw 3D points via trilinear interpolation and then used a Fully Connected Conditional Random Field (FC-CRF) to predict fine-grained class labels per point by enforcing global consistency and smoothness. Recently, \cite{graham20183d,Choy20194DNetworks} developed spatially-sparse CNNs to overcome the inefficiency of 3D-CNNs. Computational intensity, large memory used dependent on 3D voxel resolution and data loss due to sparse grid of voxelization outputs are some of the main limitations of volumetric-based methods. Therefore, these are not further investigated.

Geometric deep-learning based methods refer to the category of methods that directly apply convolutions on 3D points. These networks are chosen in our framework, since they address the challenge of point cloud data sparsity, irregularity, noise and occlusions as well as density variations \cite{agapaki2020cloi}. PointNET \cite{Qi2017PointNET:Segmentation} is a pioneering and seminal geometric deep learning network that directly extracts features per point by applying fully connected layers and concatenating the feature vectors to obtain the global features along with the class segmentation prediction scores. PointNET++ \cite{Qi2017PointNet++:Space} is one of a series of novel deep learning networks that address its limitations such as point neighborhood context and limitations of translation invariance. Multiple PointNET-based networks have been invented, each of which proposes a distinctive local aggregation operator. Some examples are PointWeb \cite{zhao2019pointweb}, RS-CNN \cite{liu2019relation}, KP-Conv \cite{thomas2019kpconv}, RandLA-Net \cite{hu2020randla}, PosPool \cite{liu2020closer}, SCF-Net \cite{fan2021scf} and PNAL \cite{ye2021learning}. Some of these networks have been applied on segmenting building objects (such as doors, windows) and structural elements (such as slabs, walls and beams) \cite{komori2019ab,wang2019graph,liang2019hierarchical,peyghambarzadeh2020point,lu2020pointngcnn,li2019tgnet,ma2020semantic}. \cite{yin2021automated} recently proposed ResPointNET++ on industrial objects (pipes, pumps, tanks, I-beams and rectangular beams), achieving 94\% overall accuracy in these classes. Their dataset is not publicly available to compare with our previous research on CLOI-NET class segmentation \cite{agapaki2020cloi}. Their results are validated on simplified industrial scenes as illustrated in Figure \ref{fig:semseglit} compared to the complex CLOI  \cite{Agapaki2019CLOI:Facilities} scenes that achieves the SOTA class segmentation performance on industrial scenes to-date (82\% overall accuracy). 

\subsection{3D Instance Segmentation}  

The performance of instance segmentation networks is dependent on class segmentation. In our previous work, we grouped the instance segmentation methods into shape-based and shape-free \cite{agapaki2020instance}. We provide a brief overview of the SOTA shape-free instance segmentation networks for completeness of this work.

3D shape-free instance segmentation methods rely on the aggregation of features per point in order to output instance labels. The features are computed either using similarity matrices for feature vector comparisons \cite{Wang2018SGPN:Segmentation,Wang2019AssociativelyClouds} or using an embedded network that measures point-wise distances \cite{Pham2019JSIS3D:Fields}. 

\cite{liu2021beacon} propose a CNN that correlates geometric and color features. However, color features have negative impacts on segmentation performance in industrial environments \cite{agapaki2020cloi}. Recently, \cite{liang2021instance} proposed the Semantic Superpoint Tree Network (SSTNet), which combines bottom-up (learning per point features) with top-down (traversing a tree of instance point clusters). However, the performance is still relatively low on benchmark datasets such as ScanNet (73.5\% mean precision and 73.4\% mean recall). Instance segmentation networks are still challenging to use in segmenting industrial scenes due to point cloud data irregularity and uncertainty on the number of instances to be segmented. Our previous work \cite{agapaki2020instance} proposed a graph-based method, which in-isolation to class segmentation (considering ground truth class labels) achieved 76.25\% mean precision and 70\% mean recall. 

The inputs of instance segmentation are the predicted point clusters from the CLOI-NET-Class method for the evaluation of the gDT framework. The data is first pre-processed in smaller 3D blocks and the outputs are point-wise instance labels. Our previous work \cite{agapaki2020instance} proposed a graph-based method, named Breadth First Search (BFS) and a boundary segmentation method to improve the BFS instance segmentation outputs. The outputs of this process are segmented point clusters that represent industrial shapes at Level of Detaol (LoD) 300. Further details on the CLOI-Ins instance segmentation method can be found in our previous work \cite{agapaki2020instance} and \cite{agapaki2020cloi}.

\begin{figure}[!ht]
\centering
\includegraphics[width=\textwidth]{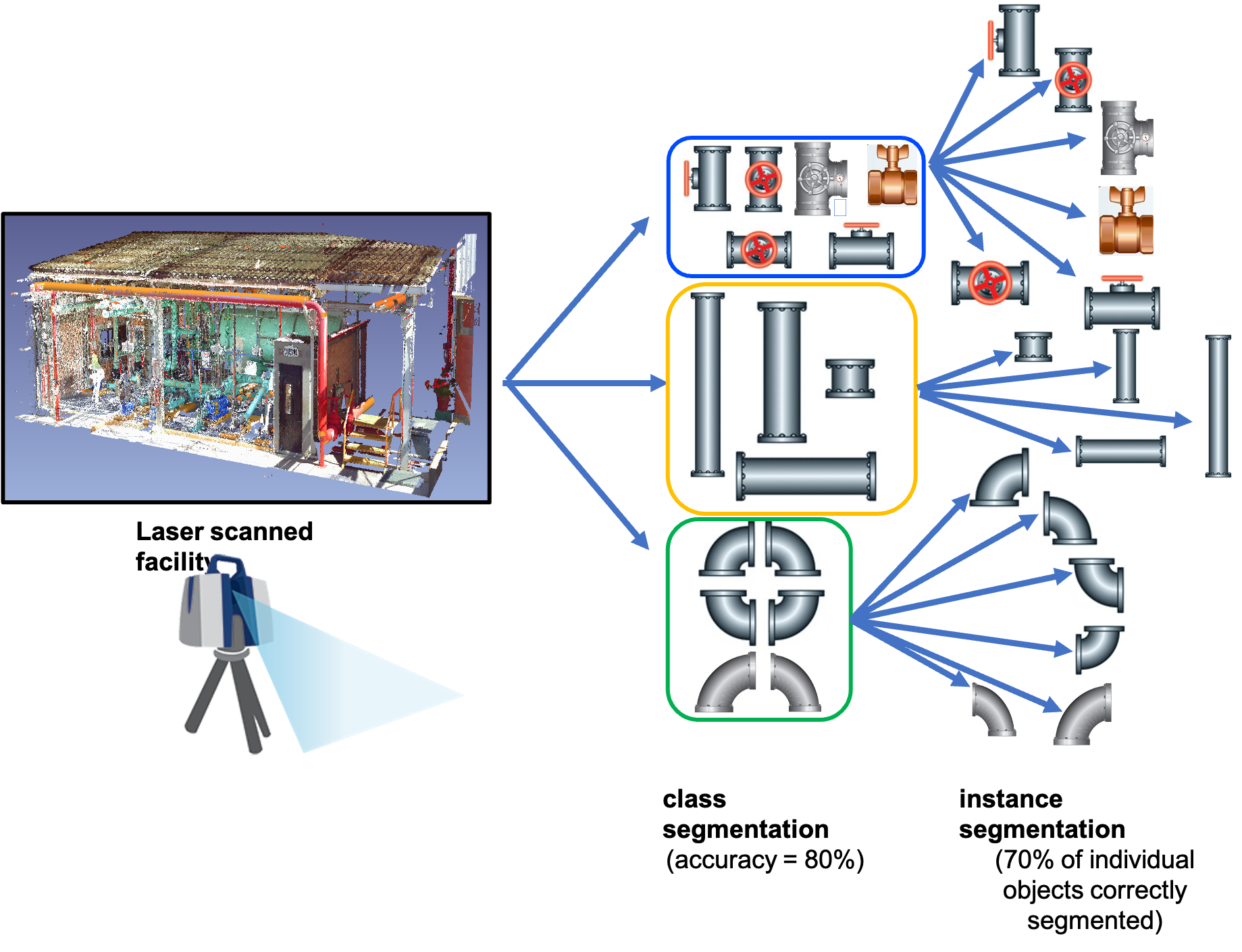}
\caption{Automated geometric Digital Twinning framework.}
\label{fig:CLOI_twinning}
\end{figure}

\begin{figure}[!ht]
\centering
\includegraphics[width=\textwidth]{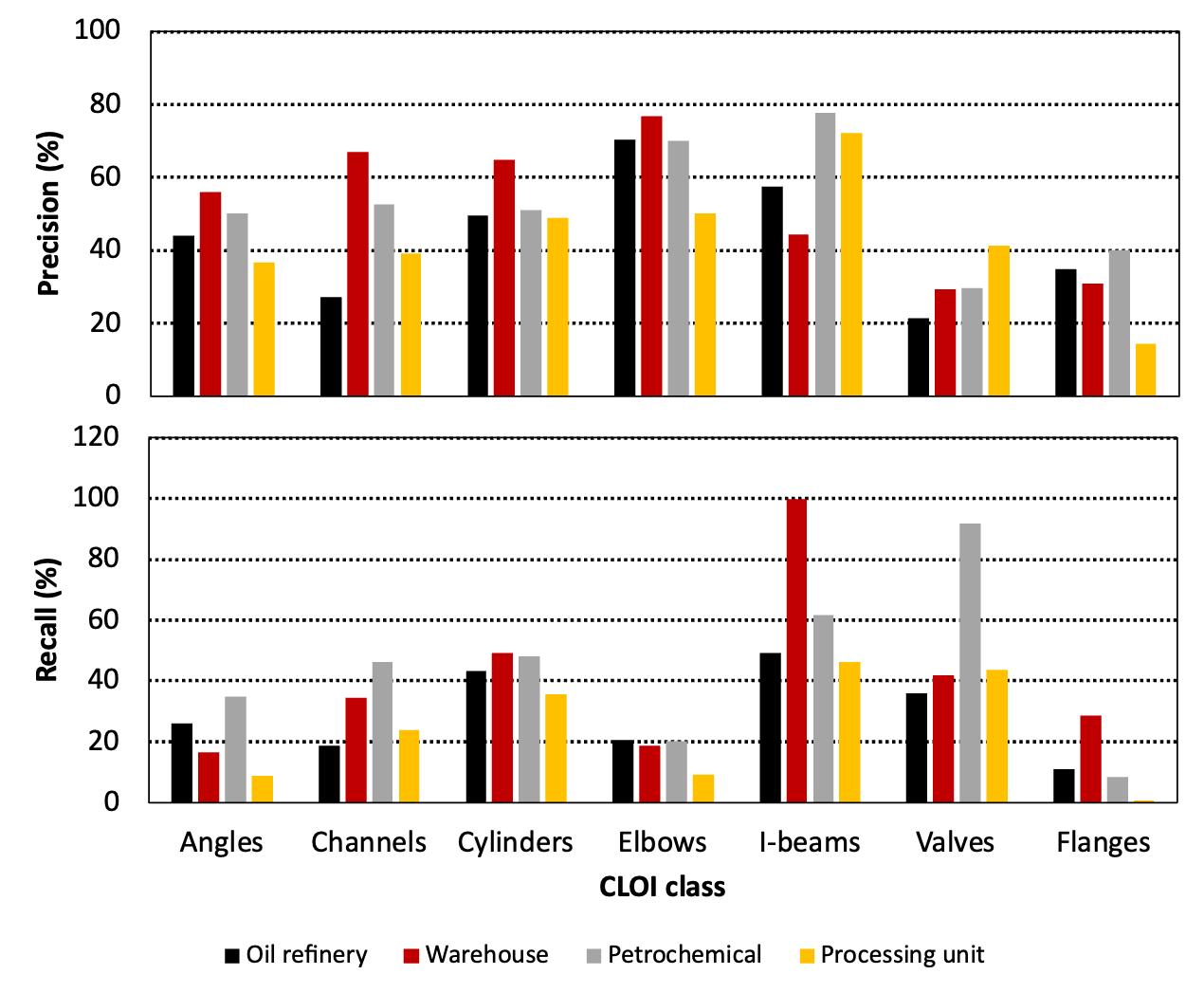}
\caption{CLOI framework performance per class for all the CLOI datasets (IoU = 25\%).}
\label{fig:CLOIPerformance}
\end{figure}

\begin{figure}[!ht]
\centering
\includegraphics[width=\textwidth]{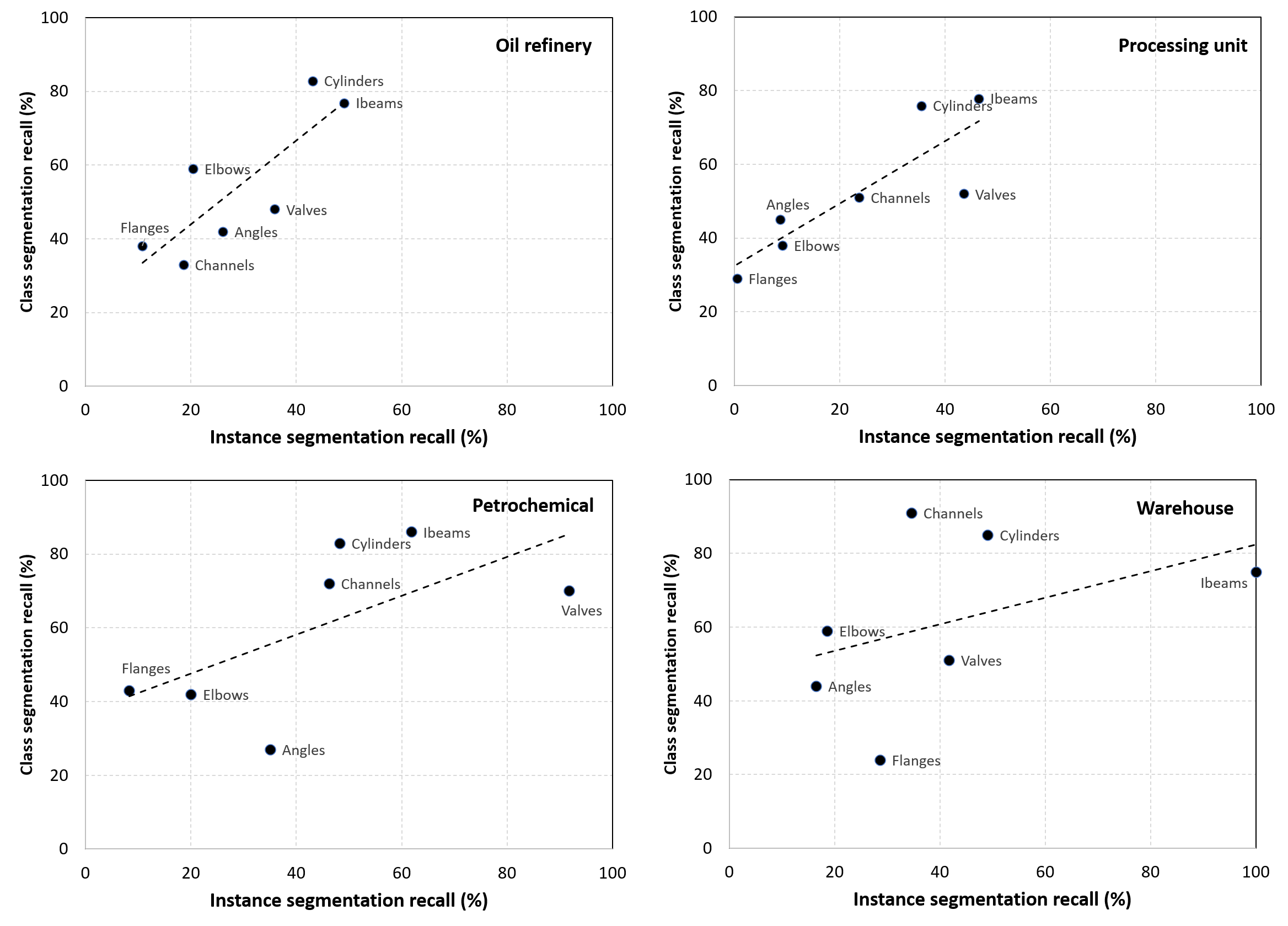}
\caption{Class and instance segmentation recall per CLOI class for the oil refinery, processing unit, petrochemical and warehouse datasets.}
\label{fig:class-ins}
\end{figure}

The CLOI automated geometric Digital Twinning framework is illustrated in Figure \ref{fig:CLOI_twinning} with the performance per task (class and instance segmentation average accuracy respectively). The average recall and precision per CLOI class and the four laser scanned datasets is summarized in Figure \ref{fig:CLOIPerformance} at Intersection Over Union $(IoU)=25\%$. The shapes with significantly higher metrics are those with higher class segmentation results. Those are cylinders (53.6\% mPrec and 44\% mRec), elbows (66.8\% mPrec) and I-beams (63\% mPrec and 64.3\% mRec). In our previous work, we investigated the impact of the low class segmentation recall on instance segmentation. There is a linear correlation between class and instance segmentation as depicted in Figure \ref{fig:class-ins}. 

Class and instance segmentation will not be further investigated, since they are out of scope of this work.

\subsection{Fitting to point clusters}

The last step of the geometric digital twinning process is fitting geometric representations to the segmented instance point clusters. This process is comprised of two main steps: (a) shape retrieval and (b) registration. The input of shape retrieval is a point cloud or mesh. The output is the CAD model that has the most similar shape with the input point cloud/mesh. Consecutively, fine registration finds the best orientation and scale of the 3D geometric representation (retrieved CAD model) by minimizing the error, which is the Euclidean distance between correspondences \cite{Salvi2007AEvaluation}.

{\bf Shape retrieval}. Shape retrieval is achieved by comparing the similarity between shapes. There are multiple shape descriptors; feature-, graph- and geometry-based \cite{tangelder2008survey}. Pipe retrieval has been extensively studied in the literature \cite{Lee2013Skeleton-basedData,masuda2015reconstruction,Liu2013CylinderPlant,Qiu2014Pipe-runClouds}, however retrieval of more complicated industrial equipment is difficult to be achieved given that these shapes cannot be parametrically defined. \cite{chen2018performance} retrieved construction objects and compared various descriptors, however retrieval accuracy was not more than 50\% in occluded objects in a synthetically generated point cloud. \cite{kim2020deep} used a deep learning based method to retrieve 3D CAD models of piping components of manually segmented industrial shapes. They used PointNET \cite{Qi2017PointNET:Segmentation} and MVCNN \cite{Su2015Multi-viewRecognition} on 4,633 pre-segmented industrial point clusters with 79.98\% and 87.41\% overall accuracy respectively. The point clusters represent piping components of the following categories: blind flange, cross, 90$^{\circ}$ elbow, non-90$^{\circ}$ elbow, flange, flange NW, olet, orifice flange, pipe, reducer CONC, reducer ECC, reducer insert, safety valve, strainer, tee, valve and wye. A method that automatically retrieves a wide range of geometrically similar industrial objects has not been achieved yet. 

{\bf Registration}. The main goal of registration after shape retrieval is to find a proper alignment for two point cloud datasets. Some of the fitting techniques are preceded by a coarse registration step, where RANSAC \cite{Chen1999RANSAC-BasedImages} or Principal Component Analysis (PCA) \cite{Chung1998RegistrationTechnique} are typically used to find the best fit of a given point cluster containing outliers. Once the datasets are roughly aligned, the fine alignment step is performed to find a more accurate fit between the point cluster and the 3D model. The majority of the methods presented for fine registration are iterative, and the associated error is monotonically decreasing. 

{\bf Point Cloud fitting}. A pioneering study that uses the closest points as the correspondences for alignment is the Iterative Closest Point (ICP) \cite{Besl1992AShapes}. ICP can be performed more robustly considering different distributions in point clouds depending on the method used for sub-sampling or selecting the control points. The control points can be sampled uniformly, randomly, or normally \cite{Rusinkiewicz2001EfficientAlgorithm} depending on the distribution of points in the point cloud dataset. Nearest-neighbour search methods also play a key role in the ICP approach. Various search types are thus used to find the nearest neighbours of the control points \cite{Greenspan2001AICP}. ICP is the most commonly used method for fine registration. Although the rate of convergence is slow compared to other methods for fine registration, its simplicity for implementation can benefit the modellers. Several variants affecting the performance of ICP are extensively investigated by \cite{Rusinkiewicz2001EfficientAlgorithm}. A novel method presented by \cite{Ziner2003AEstimation} is an example of significant improvements to ICP method. Having formulated the registration as a high dimensional optimization problem, it can be resolved using modern optimization techniques such as Genetic Algorithms \cite{Chow2002On-lineAlgorithms}. Although the presented method is faster for specific point clouds than the previously discussed registration methods, it is known as a difficult approach to be formulated and implemented. 

In the construction industry, automated registration is dominantly used for progress tracking. \cite{Nahangi2014AutomatedFabrication} presented a PCA-ICP registration for compliance checking of pipe spool assemblies. \cite{Bosche2012MarkerlessVisualization} presented a method for registration of construction laser scans with 3D/4D building models. More recently, a fully automated registration method is presented for scheduling and progress tracking \cite{Kim2013FullyMonitoring}. In the previously discussed methods, the process is semi-automated as the noise removal is being performed manually. However, Kim’s approach for automated registration includes an automated noise removal using a decay function. 

{\bf Mesh fitting}. A triangulated mesh is fitted to a point cluster by calculating the distance of a given point from the closest triangle. This method is the Iterative Closest Triangle-point ({\bf ICT}) \cite{Rabbani2004MethodsComparison} and is similar to the ICP method. The only difference is that for the ICT method, distances are calculated between points and triangles. The relative difference between distances approximated by the ICT algorithm can be significant for points that are located close to the surface. This difference is dependent on the surface type and the number of triangles that are used in the mesh. For example, the performance of ICT on curved surfaces like spheres and cylinders varies based on the number of triangles used to approximate the object.

Another option would be to convert the mesh to a point cloud and then fit the mesh generated point cloud to the point cluster. 

{\bf B-rep fitting}. There are three methods to fit B-rep models to point clusters and calculate the distance of a given point from the surface of a B-rep model. These methods work as follows:

\begin{itemize}
    \item Calculate the distance of a given point of the instance point cluster to the closest surface in the B-rep. This method is called the Iterative Closest Surface-point ({\bf ICS})
    \item Approximate the B-rep model with a triangulated mesh and calculate the distance of a given point (of the instance point cluster) from the closest triangle. This method is called Iterative Closest Triangle-point ({\bf ICT})
    \item Convert the B-rep model to a point cloud and calculate the distance of a given point to the closest point contained in the point cloud. This is the {\bf ICP} method.
\end{itemize}

B-rep fitting to point clusters has been mostly used in simple, planar building components. \cite{Valero2012AutomaticScanners} used the ICP algorithm to automatically fit B-rep models in indoor planar point clusters (walls, ceilings and floors) with high modelling precision (between 0.8 cm and 1.88 cm for vertical and horizontal edges) by intersecting the 3D planes and establishing the relationship between components. Prior knowledge is needed to fit planes using the B-rep representation, such as verticality of the walls. \cite{Valero2015SemanticTechnology} then upgraded these methods to model more objects (such as tables, chairs and wardrobes) in indoor environments by using radio frequency identification (RFID). They hypothesize that the distance of points to the fitted plane satisfies a pre-specified threshold tolerance and that the normal vector of the fitted plane and the wall point cluster is less or equal to 15$^{\circ}$. They use the ICP algorithm to fit furniture to point clusters. Their fitting algorithm has larger errors for narrow walls (above 2cm mean distance errors) and planar surfaces are considered flat, which may deviate from the existing condition of the walls. 

The reader can refer to \cite{Stroud2006BoundaryTechniques} for a comprehensive review of Boundary representation model fitting techniques.

{\bf Constructive Solid Geometry (CSG) fitting}. The CSG representation needs to be converted to one of the three previously mentioned representations (point cloud, mesh or B-rep), since direct fitting to point clusters cannot be performed.

Cuboids and other quadratic primitives are fitted to patches of points using the Glob-Fit algorithm \cite{Li2011GlobFitRelations} and CSG models are then generated \cite{Wu2018ConstructingClouds}. Glob-Fit \cite{Li2011GlobFitRelations} is a fitting method for man-made machine parts that simultaneously fits a set of geometric primitives given their global mutual relations. The geometric primitives are locally fitted using RANSAC \cite{Schnabel2007EfficientDetection} and then they are globally aligned with a constrained optimisation (i.e. extracted coaxis, coplanar, parallel/orthogonal axes). This method is dependent on correct orientation of the patches and free-form surfaces cannot be generated using this method. A disadvantage of this method is that the user should specify and fine-fune the parameters per industrial object type and computation of point normals is necessary. Another limitation of the method is that it does not consider pair of triplets (or more) primitives for fitting. Recently, \cite{Li2019SupervisedClouds} developed a supervised deep learning network that learns to fit CSG CAD industrial shapes on point cloud data, without the need for user input in terms of the parameters needed. Their Supervised Primitive Fitting Network (SPFN) is based on PointNET++ \cite{Qi2017PointNet++:Space} and learns the point normals and associated primitive type per point. The learned primitives are then matched with the ANSI 3D mechanical component library provided by the Traceparts library \cite{Traceparts2019Traceparts}.

{\bf Implicit model fitting}. Implicit model fitting to point clusters is achieved by finding the best-fitting implicit 3D representation function to a given instance point cluster, the well known Least-Squares (LS) fitting. There exist methods that optimize  the LS fitting performance. For example, \cite{Gerardo-Castro2015Laser-radarSurfaces,Gerardo-Castro2014RobustSurfaces} leverage Gaussian process implicit surfaces to reconstruct a surface with only a small subset of points \cite{Rasmussen2004GaussianLearning}. \cite{Guennebaud2007AlgebraicSurfaces} fit higher order algebraic surfaces such as spheres into a set of points that are sparsely scanned by computing the normal direction and orientation of neighboring points and applying a Moving Least Squares (MLS) algorithm. 

{\bf Free-form model fitting}. A category of methods to fit free-form models is by fitting parametric surface patches in the format of NURBS curves or surfaces. This method can be approximated as implicit patch fitting where control points determine the location of patches (NURBS surfaces). 

The formulation of the problem when fitting a free-form model is as follows: given an instance point cluster with points $(x_i,y_i,z_i) \in \mathbb{R}^3$, find a function $F: \mathbb{R}^2 \rightarrow \mathbb{R}$ that approximates the value $z_i$ at point $(x_i,y_i)$, i.e. $F(x_i,y_i) \approx z_i$. There exist many solutions to that problem which include Shepard's methods \cite{Shepard1968AData}, radial basis functions \cite{Hardy1971MultiquadricSurfaces} and finite element methods. Detailed literature reviews of these methods can be found in \cite{Schumaker1982FittingData.,Nielson1993ScatteredModeling,Franke1991ScatteredSurvey,Barnhill1977RepresentationSurfaces}.

The first step is to project the point cluster ($p_i$ points) in order to determine the $p_j$ points in $F$. This is achieved by \cite{Eck1995MultiresolutionMeshes,Floater1997ParametrizationTriangulations,Greiner1997InterpolatingB-splines}. Then, a Least Squares (LS) method needs to approximate the free-form surface for noisy point clusters. The interpolation problem can then be stated as follows: given a set of $n$ points $P_i=(x_i,y_i,z_i) \in \mathbb{R}^3$ with corresponding points $p_i=(u_i,v_i) \in \mathbb{R}^2$ and some class $S$ of parameterised surfaces $F: \mathbb{R}^2 \rightarrow \mathbb{R}^3$, find $F \in S$ such that $F(p_i) = P_i$ for all $i$. $S$ can be a bivariate polynomial, a tensor product of B-splines or NURBS, a piecewise B$\acute{e}$zier patch or a triangular patch. If $S$ is approximated by basis functions $B_1, \dots, B_k$, the interpolation function $F=\sum_{j=1}^k c_jB_j$ satisfies the function $F(p_i)=\sum_j c_jB_j(p_i)=P_i$ for all $i$. 

There are two categories of explicit free-form methods as proposed by \cite{Li2011GlobFitRelations}: (a) those that apply a smoothness prior to regularize the fitted model by favoring smooth reconstructions and (b) those that recover sharp features. The former are not suitable for standardized mechanical parts which have characteristic sharp features, unless (b) methods are then applied. 

Over the years other implicit functions have been proposed such as the Moving Least Squares (MLS). \cite{Alexa2001PointSurfaces} fit polynomial surface patches to local point neighbourhoods based on an MLS approach. This has also been used by \cite{Dey2005AnGuarantees} and \cite{Fleishman2005RobustFeatures}. A drawback of the MLS technique is that it cannot model surfaces with sharp features. \cite{Fleishman2005RobustFeatures} and methods like the power crust \cite{Amenta2001TheCrust}, the 3D Alpha Shapes method \cite{Edelsbrunner1992Three-dimensionalShapes} or the Cocone algorithm \cite{Dey2003TightReconstructor} solve this limitation.

Alternative methods have been proposed to recover sharp features called point set surface reconstruction methods. These methods are meshless and they do not exhibit any connectivity artifacts. Similar methods have been proposed using normals and MLS techniques in the literature by \cite{Fleishman2005RobustFeatures}. A limitation of these methods is that they cannot work on noisy point clouds due to the sensitivity of the MLS algorithm to errors. This limitation is addressed by \cite{Lipman2007Parameterization-freeReconstruction} who propose a Locally Optimal Projection operator (LOP) for surface approximation that is not dependent on normal estimation or fitting local planes to surface patches. 

{\bf Swept Solid fitting}. Several mathematical methods have been developed to address the sweeping of geometric entities, such as the Jacobian Rank Deficiency (JRD) method \cite{Abdel-Malek2001OnSurfaces,Abdel-Malek1997GeometricConditions}, the Sweep Differential Equation (SDE) method \cite{Blackmore1999TrimmingVolumes} and the trivariate solid visualization method \cite{Joy1992VisualizationSolids,Joy1999BoundarySolids}. General methods to determine the swept volume of parametric and B-spline entities are well-established. The JRD method has only been demonstrated in parametric and implicit sweeping with multiple parameters. However, a method by \cite{Joy1992VisualizationSolids} extended the JRD method to B-spline function to determine the implicit boundary surface. The SDE method have been demonstrated for the determination of the sweep of planar parametric curves but has not been attempted for implementing the sweeps of free-form surfaces. The reader is referred to a comprehensive review by \cite{Abdel-Malek2001OnSurfaces} addressing swept volumes, with particularly focus on the JRD, SDE and trivariate volume visualisation methods.

\begin{figure}[!ht]
\centering
\includegraphics[width=\textwidth]{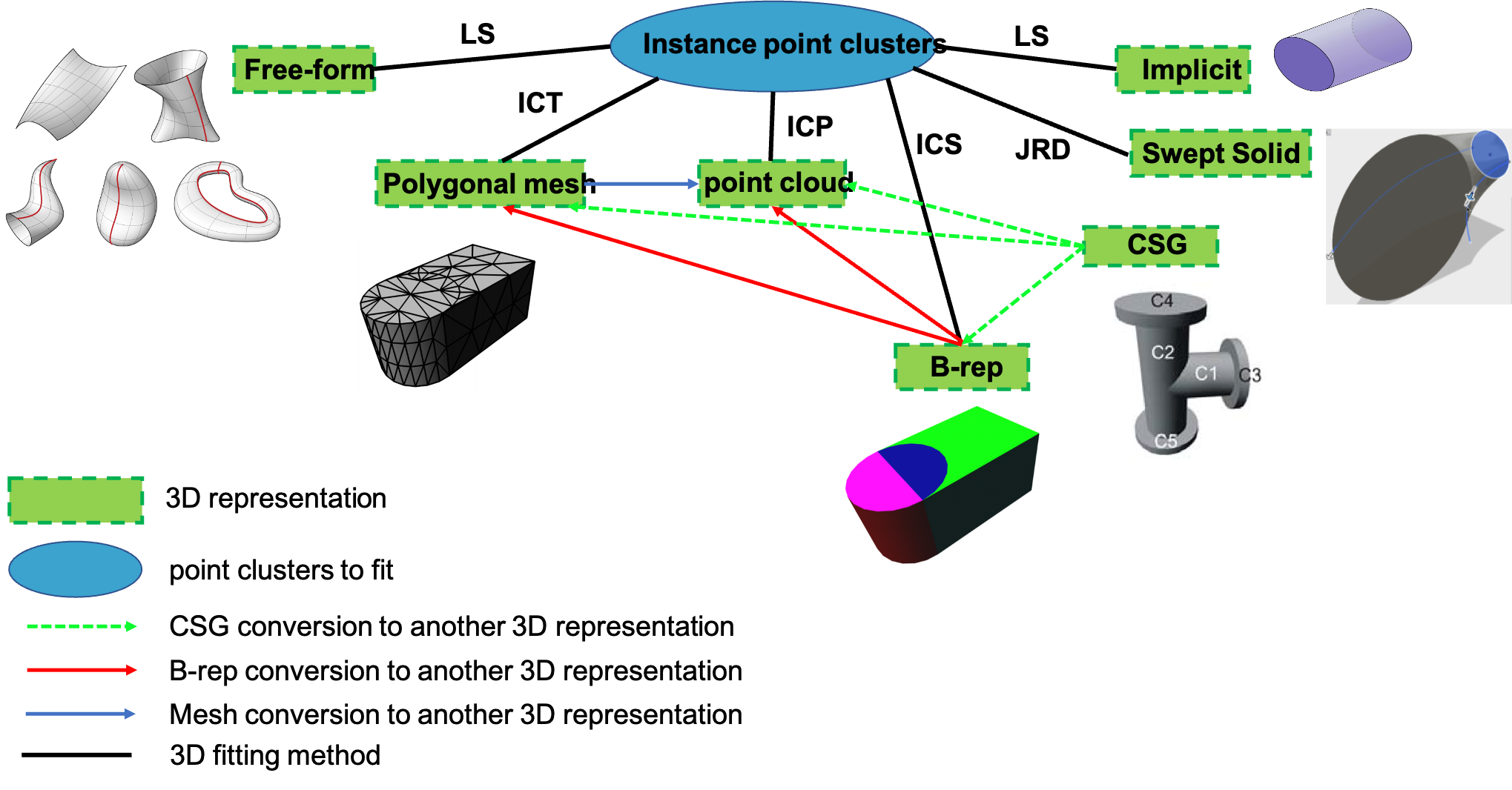}
\caption{Existing fitting methods for different 3D geometric representations and applicable conversions between 3D representations. (LS: least squares, ICT: Iterative Closest Triangle-point, ICP: Iterative Closest Point, JRD: Jacobian Rank Deficiency)}
\label{fig:fitting}
\end{figure}

This section analyzed the existing literature on 3D fitting methods from 3D geometric representations and a summary of the existing registration methods is presented in Figure \ref{fig:fitting}. The CSG method needs to be converted to either a PCD, a B-rep or polygonal mesh in order to fit instance point clusters.

It is important to note that the desired 3D representation of industrial objects may change depending on the end user application and the life cycle state of the project and the existing method is not the only solution to the fitting problem. In this case, researchers can also benefit from the advances in supervised learning methods for fitting as proposed by \cite{Li2019SupervisedClouds}. Data compression is a useful technique to perform on the instance point clusters prior to fitting that keeps only the information (based on geometry constraints) \cite{Chen2017Geometry-basedManagement}. \textit{CLOI} shapes can be represented as free-form compressed models as proposed by \cite{Lipman2007Parameterization-freeReconstruction}, however given the current state-of-practice as discussed in previous sections, swept solids can also be used depending on the application. Given the plethora of registration methods that are well-established, further investigation on those will not be performed in this research.

\subsection{Gaps in Knowledge, Objectives and Research Questions (RQs)}

Considering the SOTA practice and body of research reviewed above, our previous research on class and instance segmentation \cite{agapaki2020cloi,agapaki2020instance} generated instance point clusters at a commercially viable level. However, 3D CAD models that correspond to the CLOI gDT instance point clusters have not been retrieved. This is a significant gap in knowledge (GiK) that prevents the effective generation of a gDT. Previous methods of CAD shape retrieval on point clouds have the following limitations: (GiK 1) the instances are pre-segmented incurring significant amount of manual labor and (GiK 2) the methods have not been tested on a benchmark dataset with multiple industrial classes with close similarity. They were only tested on 18 distinctive piping components \cite{kim2020deep}.

Out of these GiK, we seek to address GiK 2 by answering the following Research Questions:

\begin{itemize}
    \item{\emph{RQ1}:} How to automatically retrieve the most similar CAD model to segmented instance point clusters? 
    \item{\emph{RQ2}:} How sensitive is the shape retrieval to the quality of laser scanned data (pre-segmented instance point clusters)?
\end{itemize}

\section{Proposed solution}
\label{proposed}

We target to solve the problem of 3D shape retrieval of segmented industrial point clusters with respect to shape similarity. The primary objective of this paper is to derive a benchmark methodology that can first be used in the shape retrieval of ``ideal'' instance point clusters with the potential to be the foundation for future research in the shape retrieval of segmented instances. This objective will be achieved by answering the above-mentioned research questions.

\subsection{Assumptions}

We assume that the proposed 3D shape retrieval method is feasible in the context of gDT generation under the following assumptions:

\begin{enumerate}\setcounter{enumi}{0}
  \item [{\bf A1.}] The input data of the proposed method is ideal instance point clusters. Noisy points and outliers were manually removed. 
  \item [{\bf A2.}] A 3D CAD model is available for each instance point cluster. 
  \item [{\bf A3.}] Partially scanned or occluded instance point clusters are excluded from this analysis.
\end{enumerate}

\begin{figure}[!ht]
\centering
\includegraphics[width=0.65\textwidth]{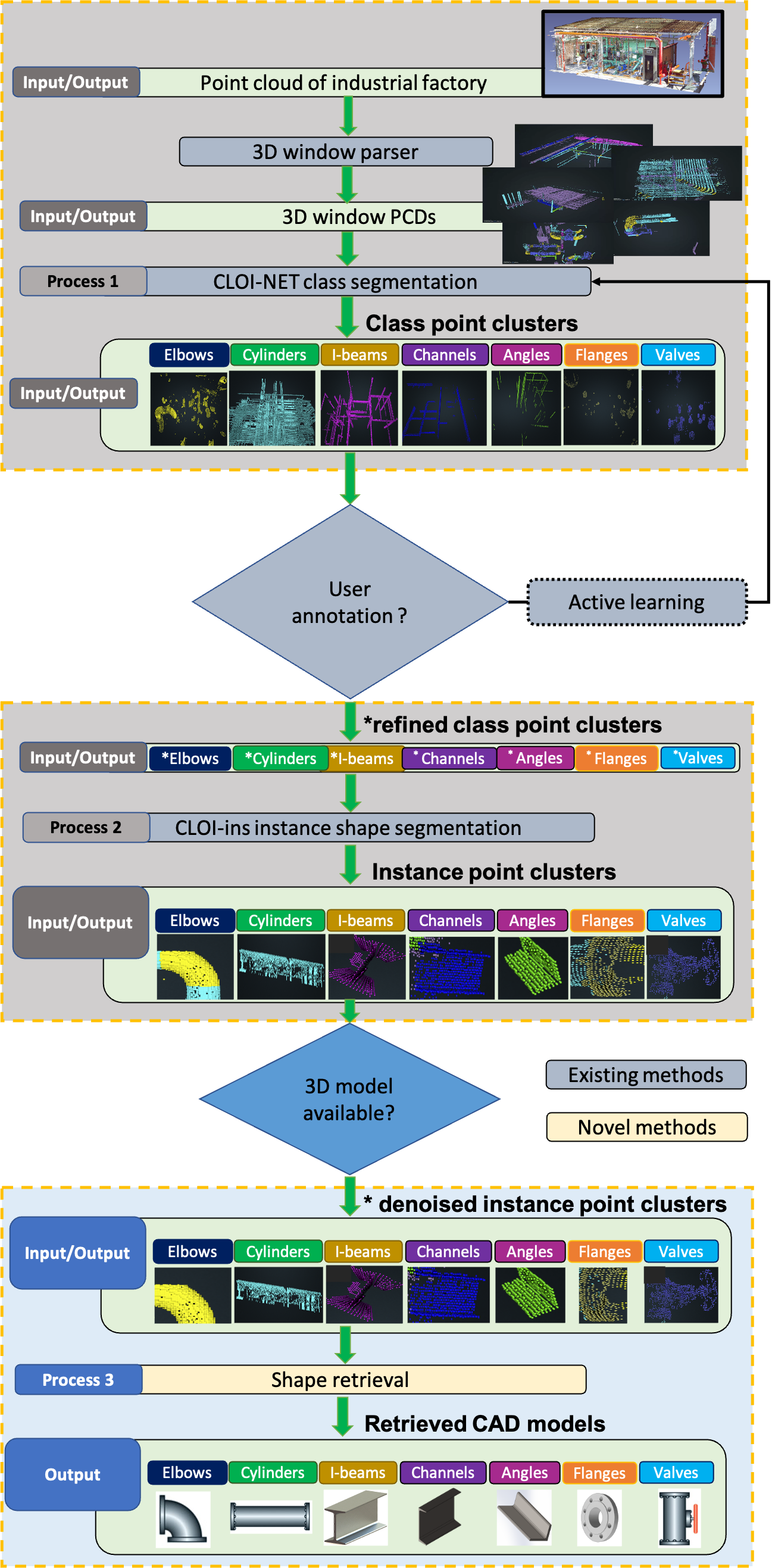}
\caption{Workflow of the proposed \textit{CLOI} framework.}
\label{fig:framework}
\end{figure}

The proposed framework consists of three major parts. Specifically, these parts are {\bf (1) class segmentation} \cite{agapaki2020cloi}, {\bf (2) instance segmentation} \cite{agapaki2020instance} and {\bf (3) shape retrieval} as presented in Figure \ref{fig:framework}. Processes 1 and 2 were addressed in our previous work as described in the Background section. This research focuses on Process 3. The inputs of our novel proposed method are manually segmented instance point clusters. These ``perfectly`` segmented instances are used to evaluate the accuracy of the method in isolation without adding the errors of the class and instance segmentation predictions from Processes 1 and 2.

\subsection{Overview}

\begin{figure}[!ht]
\centering
\includegraphics[width=\textwidth]{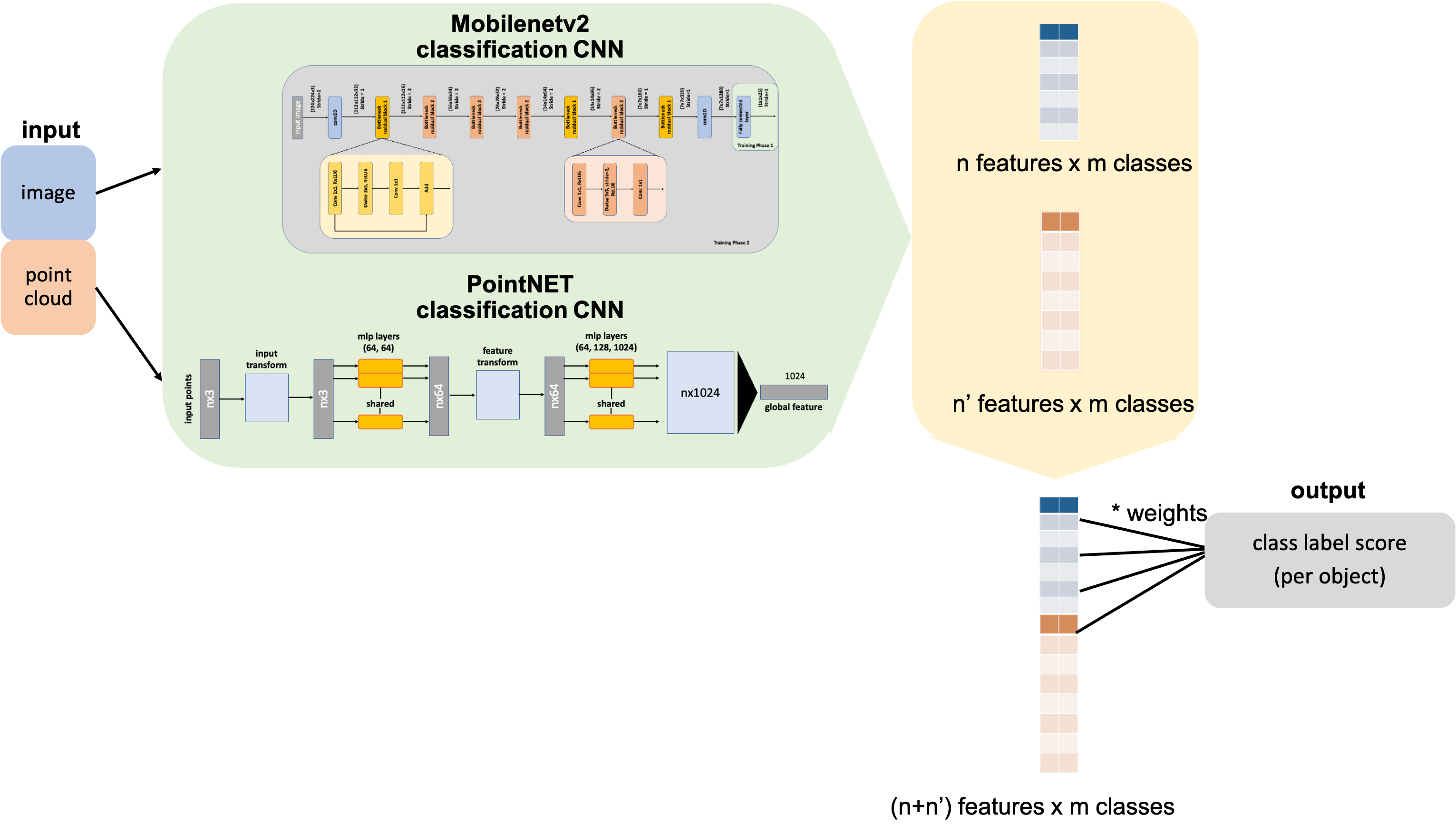}
\caption{Workflow of the proposed Joint classification network.}
\label{fig:joint}
\end{figure}

We aim to extract features of the segmented instances and use those to identify the most similar 3D CAD model. Figure \ref{fig:joint} illustrates the proposed method that consists of two main steps. {\bf Step 1} is the instance classification network, where the feature vector of each instance point cluster will be extracted. The inputs of this method are the instance point clusters and the outputs are a feature vector per instance. {\bf Step 2} is the similarity-based shape retrieval, where each instance feature vector is compared to the feature vectors of all 3D CAD models. A similarity metric is defined to measure the distances between the distributions of the feature vectors. The 3D CAD model with the smallest distance to the instance point cluster is the correctly identified shape and output of the method. The following sections expand on each step of the method in detail.

The selection of the method is based on experimental evaluation of each step of the SOTA shape retrieval deep learning networks that are suitable for the identification of industrial components. For example, some criteria that the methods need to address are: (1) the complexity of industrial object geometry (e.g., curved or multi-faceted surfaces of elbows and valves) and (2) slight geometric differences between shapes (e.g., hex nut and lock nut). 

The methods that address the above-mentioned criteria are geometric deep learning-based methods and the best-performing networks are PointNET \cite{Qi2017PointNET:Segmentation}, PointNET++ \cite{Qi2017PointNet++:Space} and their derivatives. Therefore, we first investigate the performance of these networks on available industrial/mechanical part benchmark datasets. 

\subsection{Step 1 - Instance Classification}

We first investigate the performance of SOTA point cloud and image networks as well as available commercial software on benchmark datasets. The largest annotated benchmark of mechanical parts for classification and retrieval tasks is the Mechanical Components Benchmark (MCB) \cite{sangpil2020large} with two datasets: MCB (A) with 68 mechanical classes of industrial objects and 58,696 CAD models and MCB (B) with 25 classes and 18,038 CAD models. These objects were retrieved from online CAD catalogs (TraceParts, 3D Warehouse and GrabCAD). We selected MCB (B) for our evaluation as a better proxy to real industrial objects that are not consistently oriented. That will help the generalization of shape descriptors \cite{savva2016shrec16}. Table \ref{table:benchmark} shows the object distribution across the 25 classes of the MCB (B) dataset that is used in our validation. It is noteworthy to mention that the reason for selecting the MCB (B) dataset as opposed to the CLOI dataset are the diversity of industrial classes (e.g., there are many geometrically similar objects that belong to separate classes) and the availability of CAD models that are needed for feature extraction.

We then validate two deep learning networks: a view-based method that outputs the class label of 3D models by projecting them into 2D images and a point cloud method that converts the 3D models into point clouds, directly processes them and outputs the class label. We selected MobileNetV2 \cite{sandler2018mobilenetv2} and PointNET \cite{Qi2017PointNET:Segmentation} as the backbone networks of our analysis. We also used the Microsoft Custom Vision (MSFT CV) service by Microsoft Azure \cite{Azure2022} given the large pre-trained library of image classifiers they have. MSFT CV has a limitation on the number of images for training (100,000 images). Therefore, we only rendered 4 images per 3D model to be consistent with our comparison. We used the same split as defined in the MCB (B) benchmark dataset (80\% train and 20\% test).

\begin{figure}[!ht]
\centering
\includegraphics[width=\textwidth]{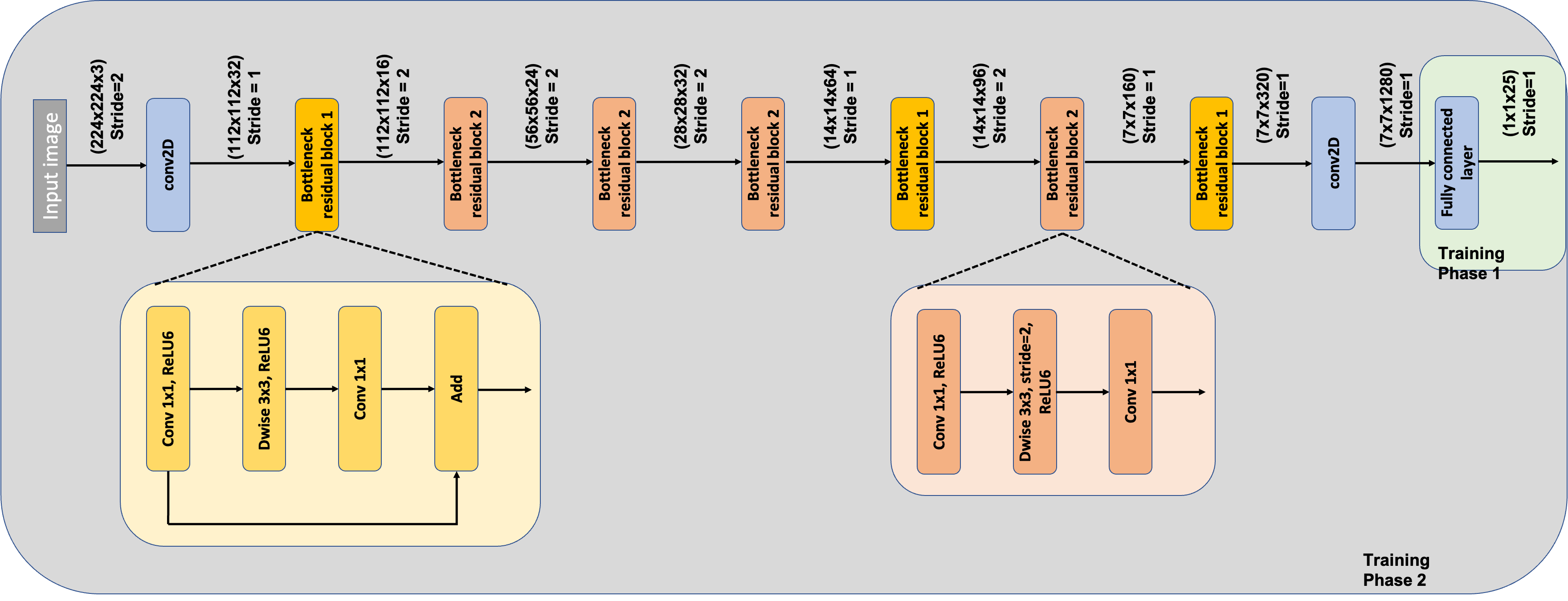}
\caption{Architecture of MobileNETv2 classification network and training phases.}
\label{fig:mobilenetv2}
\end{figure}

We selected MobileNETv2 to benchmark our proposed method given its lightweight architecture and accuracy on existing benchmark classification datasets (e.g., ModelNET40 \cite{wu20153d}). Figure \ref{fig:mobilenetv2} summarizes the MobileNETv2 architecture and the two training phases. In training phase 1, only the last layer of the network is used for training (fully-connected layer) for 20 epochs and training phase 2 uses the entire network for training (40 epochs). In other words, in the first training phase we freeze all the layers except the fully connected layer and then repeat training the entire network. This training strategy is used to account for fine-tuning the network parameters as suggested by \cite{Tensorflow}. 

Each image is processed using this architecture and the output is a probability score for each CAD model of the 25 MCB(B) classes. We then use the product of the probability distributions of the 4 rendered images that represent the CAD model to get the output probability distribution and then the class label score. Similarly, we use the PointNET \cite{Qi2017PointNET:Segmentation} architecture as suggested by \cite{kim2020deep} for instance classification.

{\bf Image augmentation parameters}. The input size of the 4 rendered images is 224x224. The images were rendered using the Panda3D library \cite{panda3d}. We use standard image augmentation techniques to rotate (rotation range=20$^{\circ}$), shift (width and height shift range=0.2$^{\circ}$) and zoom (zoom range = 0.2$^{\circ}$) the images while training. Image rotation is needed to account for orientation invariance of the object to be classified. Random shift is another important image augmentation parameter that addresses the position of the object in the image (e.g., the object may not always be in the center of the image). Random flip is also introduced to flip the object along the vertical and horizontal axis. Random zoom is considered to randomly zoom in on/out of the image.

{\bf Point cloud rendering parameters}. We use the following parameters for rendering point clouds from the MCB (B) 3D CAD models. The CAD models are in mesh format (.OBJ files), which we first convert into point clouds, using the Trimesh library \cite{trimesh}. We uniformly and randomly sample 2,048 points from each mesh; we selected this number after experimenting with 4,096, 8,192 and 16,384 points. Increasing the number of points did not significantly affect the average accuracy (it only increased by 2-4\%), consistent with the experiments of \cite{Qi2017PointNET:Segmentation} on benchmark point cloud classification tasks.

\begin{figure}[!ht]
\centering
\includegraphics[width=1.1\textwidth]{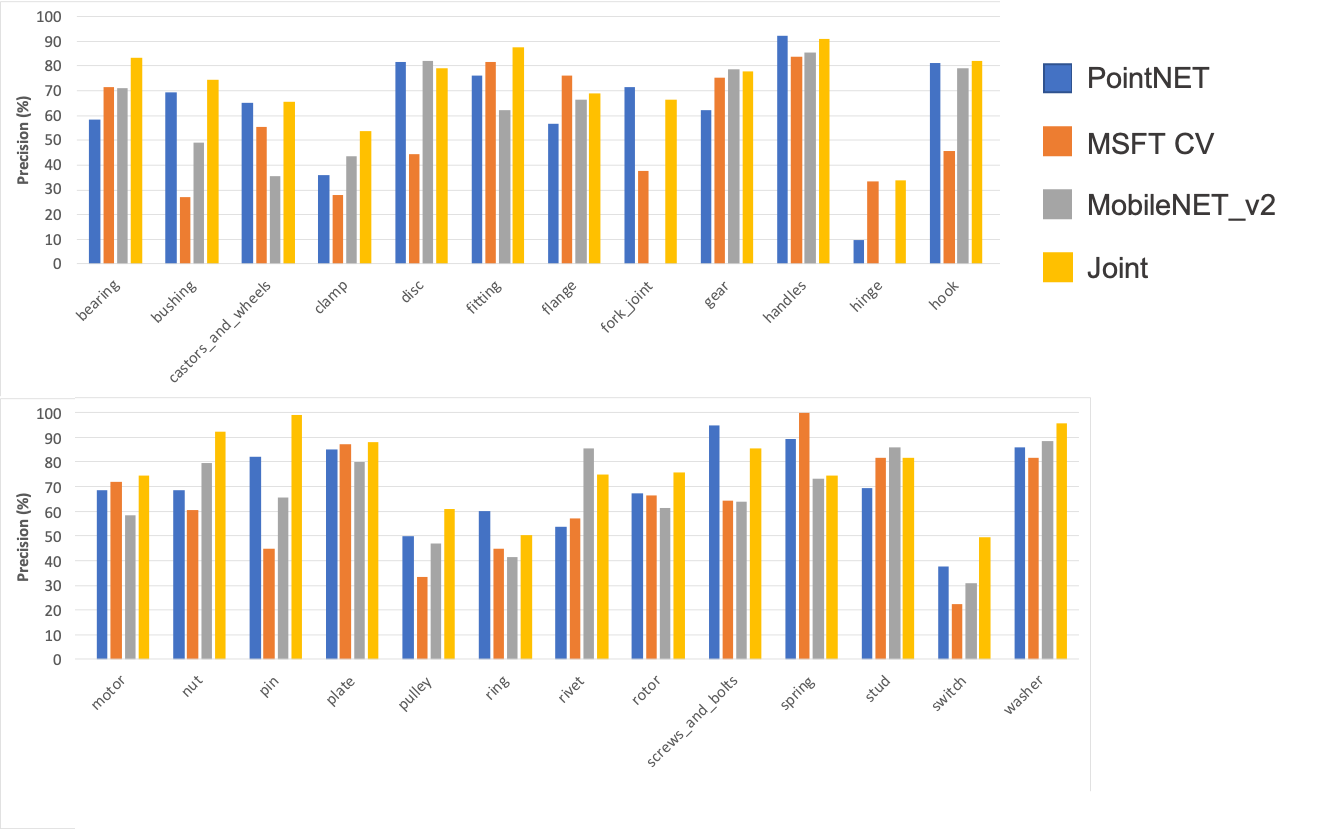}
\caption{Per class classification precision on the MCB(B) dataset and comparisons with PointNET, MobileNETv2 and Microsoft Custom Vision (MSFT CV).}
\label{fig:jointprecision}
\end{figure}

\begin{figure}[!ht]
\centering
\includegraphics[width=1.1\textwidth]{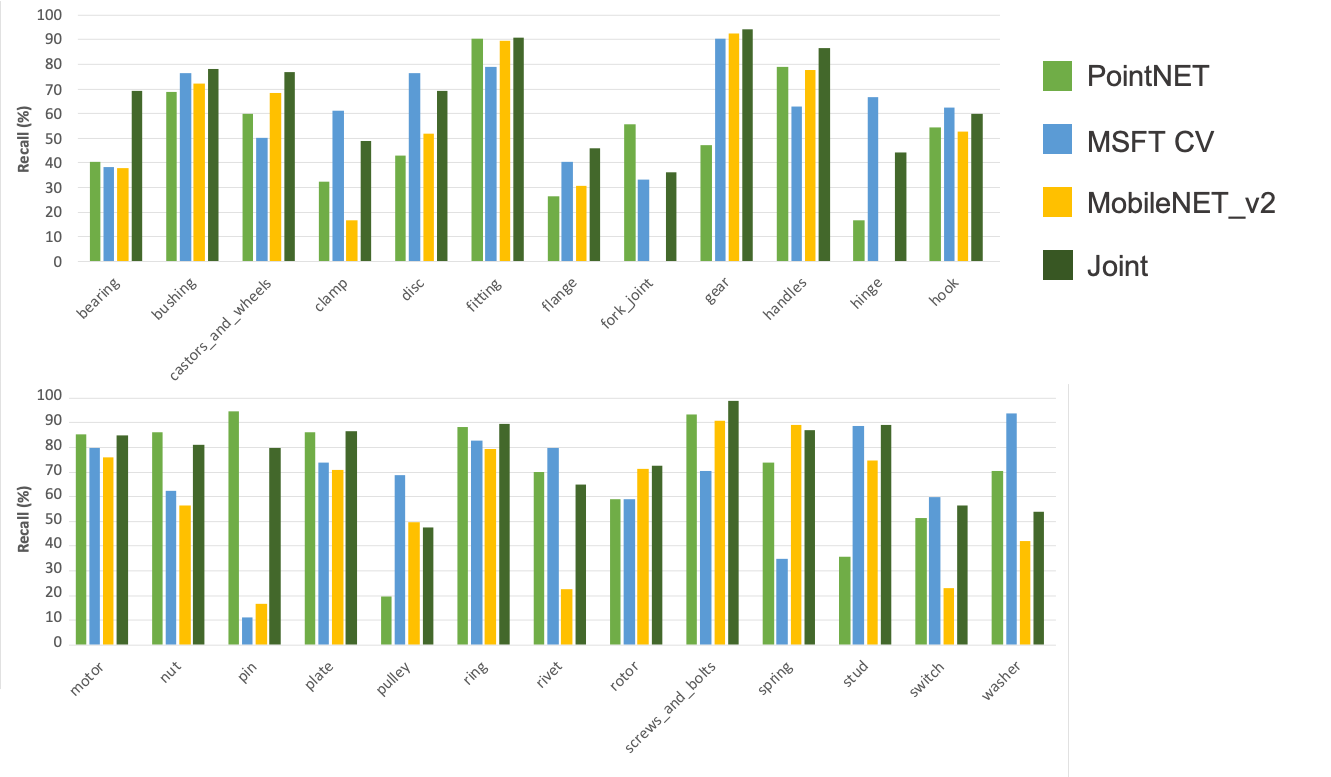}
\caption{Per class Classification recall on the MCB(B) dataset and comparisons with PointNET, MobileNETv2 and Microsoft Custom Vision (MSFT CV).}
\label{fig:jointrecall}
\end{figure}

Table \ref{table:classificationnetworks} summarizes the performance of the two selected networks, showing that PointNET has slightly superior performance on the MCB (B) dataset compared to MobileNETv2. The metrics of the PointNET network are 66.9\% precision, 61.1\% recall and 63\% accuracy on average as opposed to 60.6\% precision, 54.1\% recall and 64\% accuracy of MobileNETv2. The classification precision and recall per class are illustrated in Figures \ref{fig:jointprecision} and \ref{fig:jointrecall} for MobileNETv2, PointNET and MSFT CV respectively. We observe that PointNET has consistently better performance compared to MSFT CV and MobileNETv2, however it is not clear whether one of the two outperforms in all classes. For example, the precision of the screws and bolts class is higher when using PointNET (94.74\%) compared to using MobileNETv2 (62.75\%) for classification. However, there is a reverse trend for other classes such as studs. The precision of studs is higher when using MobileNETv2 (85.83\%) compared to using PointNET (68.83\%). Therefore, we infer that both visual features from images and geometric features from point clouds can be significant for correctly classifying complex mechanical parts.

Recently, the use of multiple CNNs jointly as feature extractors has improved image clustering by 10\% \cite{guerin2018improving}. Given that we can generate both images and point clouds from CAD models, point cloud data can be processed in parallel with the images. In our proposed instance classification network, we investigate whether the use of multiple CNNs can recognize useful complementary features when trained on the same data, therefore improving the classification performance. In particular, the idea is to develop a joint deep learning network that uses a deep CNN for processing images and another network for processing point cloud data. The components of the joint network (shown in Figure \ref{fig:joint}) are trained jointly in the same way that each separate network would be trained as described above.

\begin{figure}[!ht]
\centering
\includegraphics[width=1.1\textwidth]{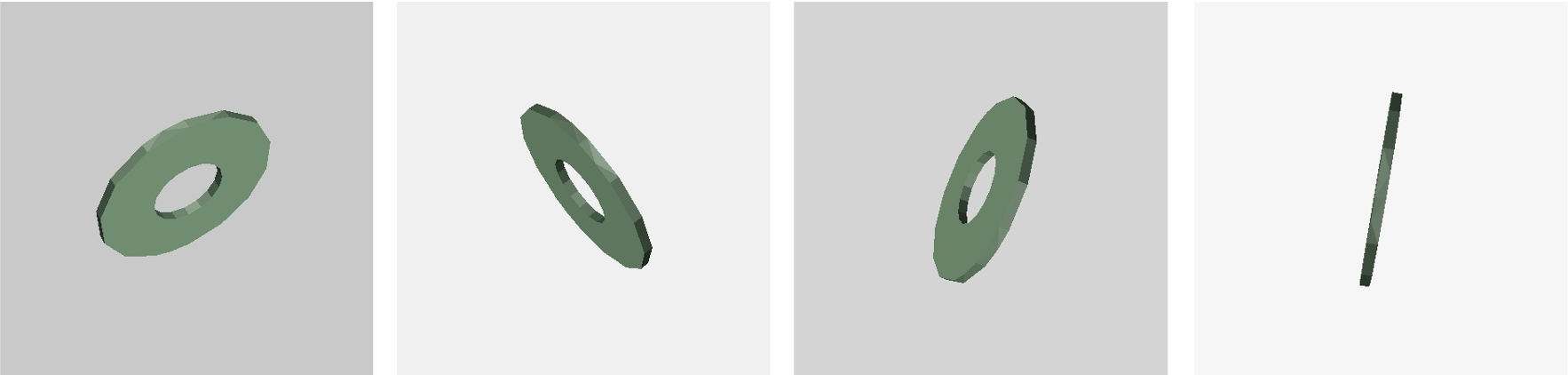}
\caption{Rendered images of a washer using Panda3D.}
\label{fig:panda3drender}
\end{figure}

We select the ModelNETv2 and PointNET networks based on the evaluation of image and point cloud classification networks as discussed above. Therefore, the feature vector of the ModelNETv2 network (with $n\ features\ x\ m\ classes$) and the feature vector of the PointNET network (with $n'\ features\ x\ m\ classes$) are concatenated resulting in the joint feature vector (with $n+n'\ features\ x\ m\ classes$). The output (probability scores and class labels per CAD model) are computed based on the concatenated feature vector. The performance of the joint network is illustrated in Figures \ref{fig:jointprecision} and \ref{fig:jointrecall}, where it outperforms MobileNETv2, PointNET and MSFT CV. There are a few exceptions such as the ``screws and bolts'' class, where PointNET network in isolation has higher precision (94.74\%) or the ``washer'' class, where MSFT CV has the highest recall (93.95\%). The rendered images of a ``washer'' are presented in Figure \ref{fig:panda3drender} and we observe that a reason of the low recall performance of MobileNETv2 network on that class (42.74\%) can be attributed to the fact that one of the images is not representative of a washer. We intend to increase the number of views to avoid that issue in future work.  

\subsection{Step 2 - Similarity-based Shape Retrieval}

The feature vector of each 3D CAD model is then compared with the feature vector of the segmented point cloud instance to identify the most similar CAD model. The simplest distance we propose to use is Euclidean distance. 

\begin{figure}[!ht]
\centering
\includegraphics[width=1.1\textwidth]{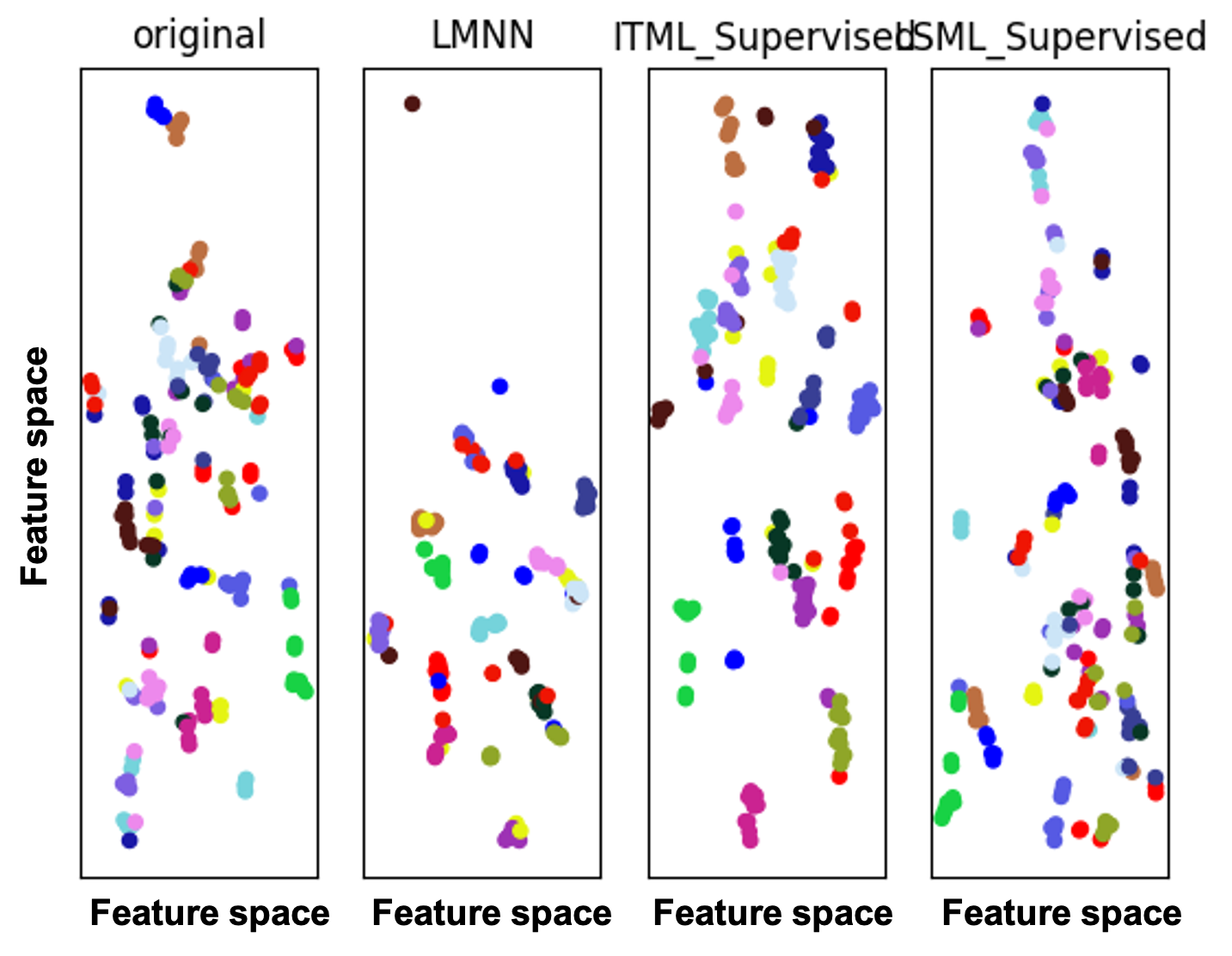}
\caption{Comparison of clustering methods on the MCB-A partial dataset (only the objects that are not included in training). Each color represents a distinct cluster.}
\label{fig:metriclearning}
\end{figure}

We also investigated the following widely used feature vector distances: (a) Large Margin Nearest Neighbor (LMNN) \cite{weinberger2009distance}, (b) Information Theoretic Metric Learning (ITML) \cite{10.1145/1273496.1273523} and (3) Least Squares Metric Learning (LSML) \cite{liu2012metric}. The distance between the feature vector distributions in those methods is the Mahalanobis distance \cite{mahalanobis1936generalized}. We used the joint network described in {\bf Step 1} pre-trained on the 25 mechanical classes of the MCB-B dataset (with the network parameters as specified above) and retrieved the feature vectors per shape. Then, we used the three methods identified above to cluster mechanical objects of the MCB-A dataset that are not present in the MCB-B dataset (test set). The results are presented in Figure \ref{fig:metriclearning}. We observe that objects of the same class (which are represented in the same color in the Figure) are not clustered properly; some shapes of the same class are grouped in the same class based on their distance in the feature space. Therefore we did not consider those methods for further analysis. Each shape is represented with a colored circle in the Figure. The features represented in the figure are computed after running a Principal Component Analysis (PCA).

In our previous work, we developed a gDT generation framework with instance point clusters as outputs. This work is the next step of the gDT generation process and an essential step of fitting geometric shapes to the extracted instance point clusters. The novelties of this work are (1) evaluating a new instance classification method that identifies the geometric features of instance point clusters and (2) proving that it fetches their counterpart 3D CAD models from existing CAD catalogs or BIM models.

\section{EXPERIMENTS AND RESULTS}
\label{methodology}

\subsection{Implementation}

\begin{figure}[!ht]
\centering
\includegraphics[width=\textwidth]{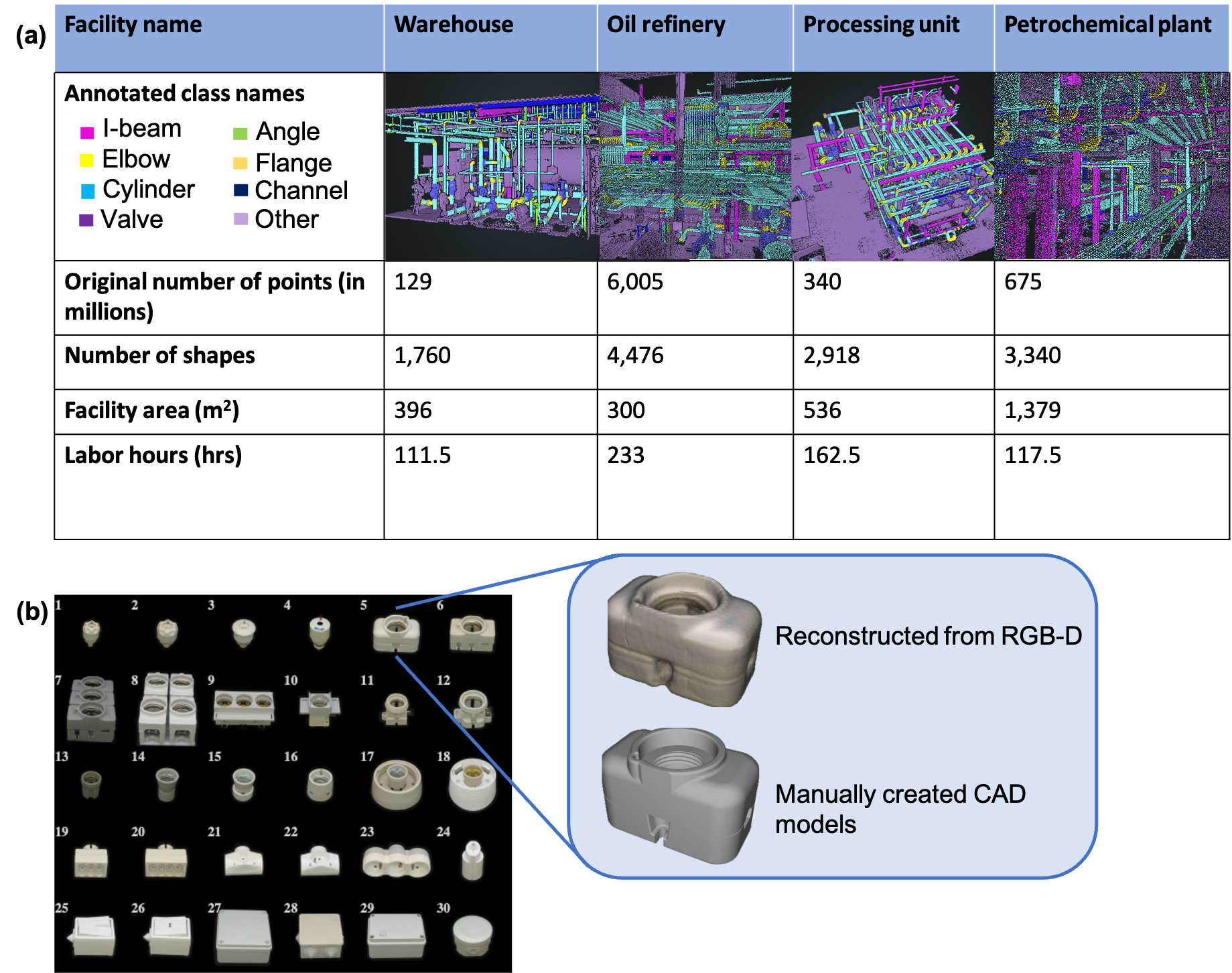}
\caption{(a) CLOI benchmark and (b) T-LESS dataset specifications.}
\label{fig:CLOI_benchmark}
\end{figure}

We validate the shape retrieval methodology on the T-LESS dataset, which has both 3D CAD models and reconstructed objects. This was the only dataset of industrial parts that complies with the data requirements as identified in our proposed solution above. Figure \ref{fig:CLOI_benchmark} shows the T-LESS dataset \cite{hodan2017t}, which is composed of 33 shapes. The reconstructed dataset was generated using three synchronized data capturing sensors; a Primesense CARMINE 1.09 (structured-light RGB-D sensor), Microsoft Kinect v2 (a time-of-flight RGB-D sensor) and a Canon IXUS 950 IS (a high-resolution RGB camera). The dataset is composed of thirty commodity electrical parts that have no significant texture, discriminative color or distinctive reflectance properties and often have similar shape and/or size. The CLOI dataset could not be used for validating our proposed methodology given that no 3D CAD models are available for the dataset.

We developed a proof of concept prototype and implemented the methodology on Tensorflow 2.0. We used a Google Cloud VM with NVIDIA Tesla P100 GPUs to run our experiments.

\subsection{Image rendering}

A prototype was developed using the Panda3D game engine to render photorealistic 2D images of the T-LESS CAD models. The camera is moving on the horizontal plane around the center of the object. The image size used was 224x224, the background was set to random light and scale to 0.5. The rendering orientation of the object is set to heading$=0$, pitch=$90^{\circ}$ and roll$=0$. This is frequently necessary when the object is oriented upside down, so that we sample it properly with a camera moving on the hemisphere.

\begin{figure}[!ht]
\centering
\includegraphics[width=1.1\textwidth]{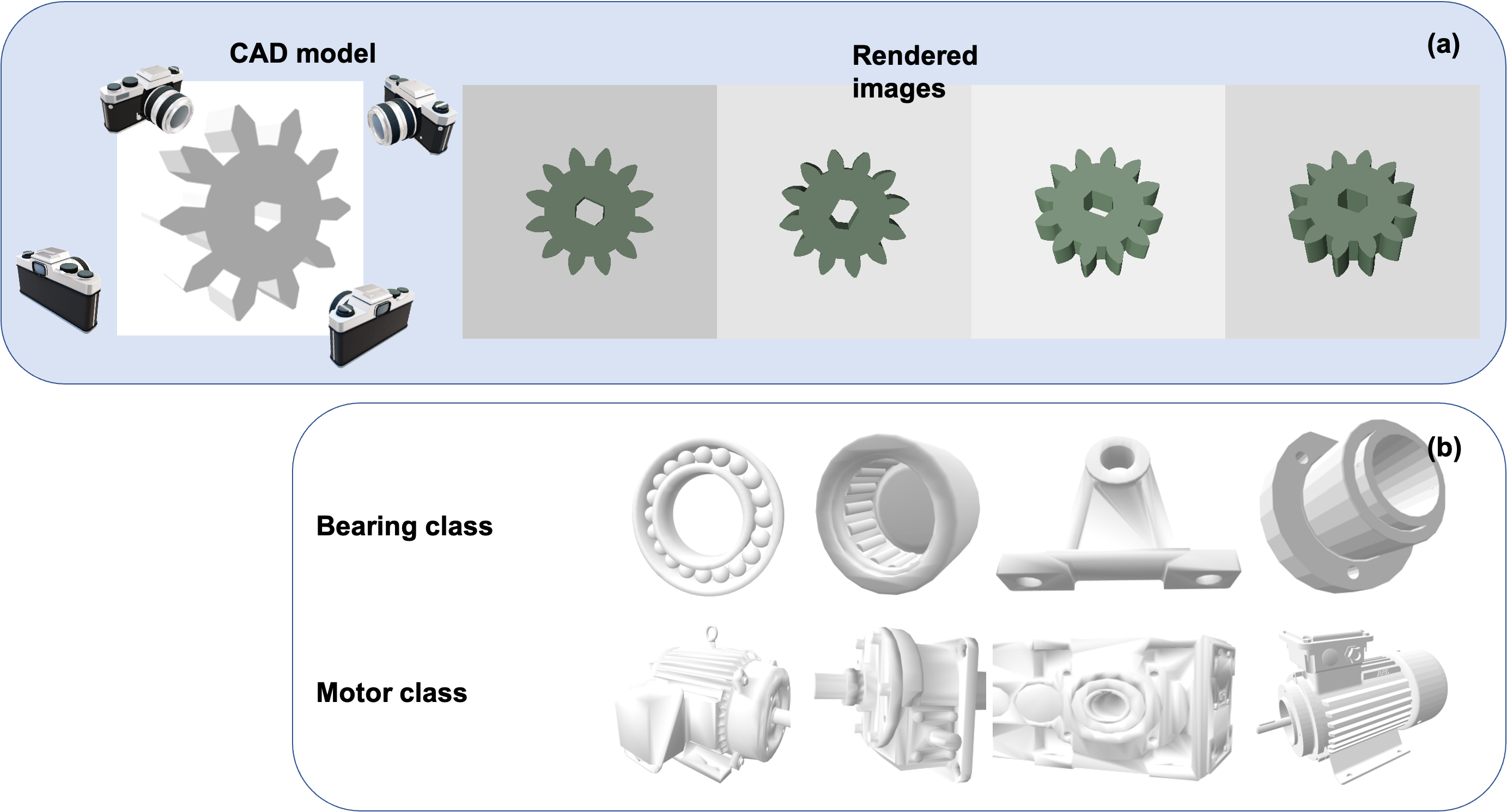}
\caption{(a) CAD model of a gear from the MCB(B) dataset and rendered images from Panda3D prototype and (b) CAD models of the bearing and motor classes.}
\label{fig:rendering}
\end{figure}

Some examples of generated images and their respective CAD models are illustrated in Figure \ref{fig:rendering}(a). Each class has high intra-class variations as shown in Figure \ref{fig:rendering}(b). This validates the case where class and instance segmentation processes are not sufficient to extract all the relevant information for industrial objects if the shape retrieval process is not performed.

An assumption of the developed prototype is that textures are not considered in the rendered images and CAD models.

\subsection{Point cloud rendering}

A prototype is developed using the Trimesh library \cite{trimesh} with the configuration and parameters analyzed in the proposed solution.

\subsection{Experiments and Discussion}

\begin{figure}[!ht]
\centering
\includegraphics[width=\textwidth]{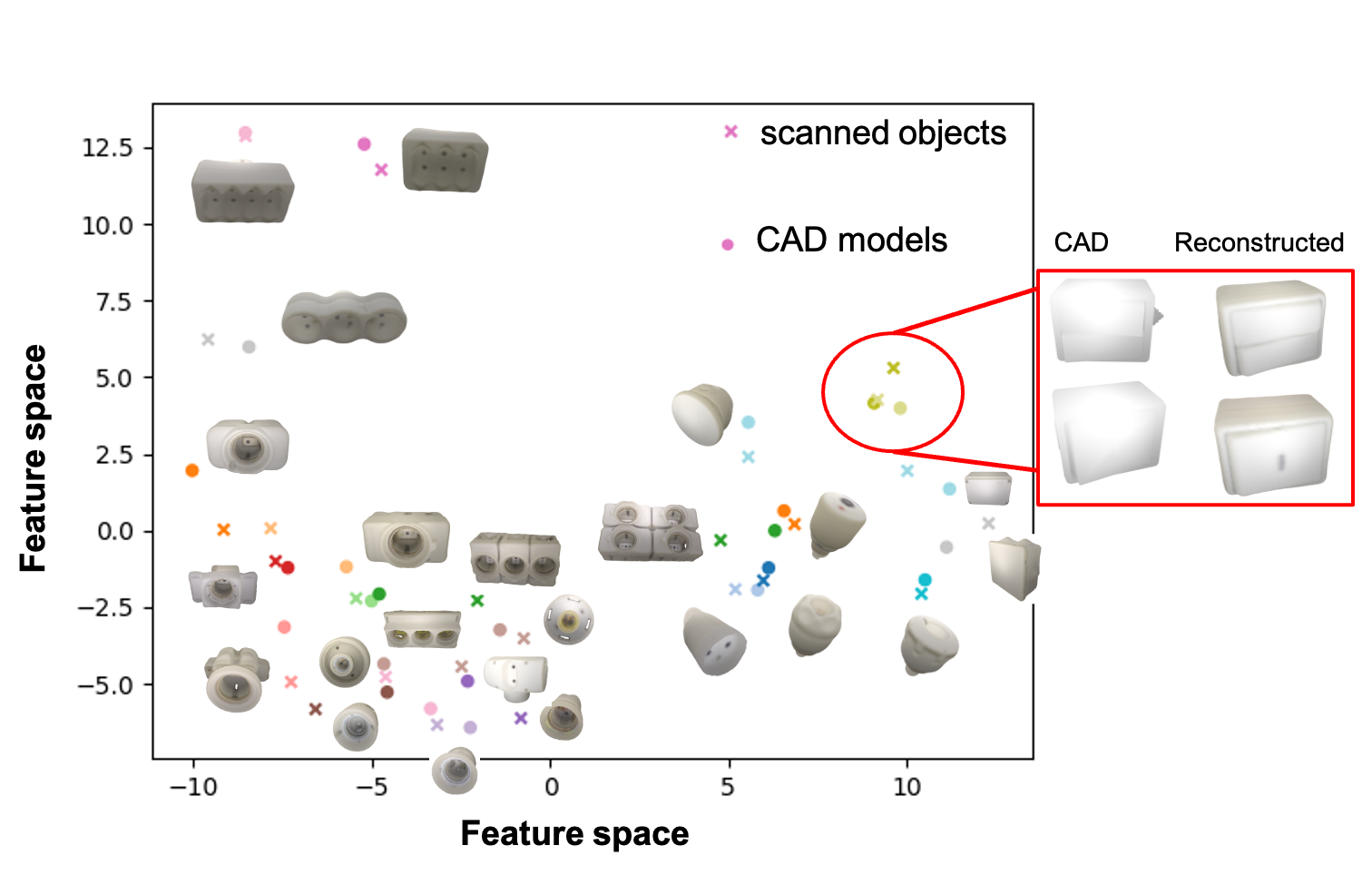}
\caption{Principal Component Analysis (PCA) Visualization of feature vectors for CAD and reconstructed models of the T-LESS dataset.}
\label{fig:pca}
\end{figure}

The performance of the shape retrieval methodology was evaluated based on the performance on the T-LESS dataset. The inputs of the proposed method are the images and point clouds of the T-LESS electrical components derived using the prototypes above. The first step is to validate the instance classification method on the T-LESS dataset. Therefore, the joint-network is pre-trained on the MCB-B dataset (both train and test sets) and then tested on the T-LESS dataset. The concatenated feature vector is then taken for each CAD and reconstructed model of the T-LESS dataset and plotted in feature space. Figure \ref{fig:pca} presents the 2D representation of the feature embedding. The dimensions are reduced to 2D using Principal Component Analysis (PCA). This figure showcases that only two out of the 30 T-LESS objects are not retrieved. Table \ref{table:classificationnetworks} summarizes the top 1, top 3 and top 5 accuracy being 85.2\%, 88.9\% and 88.9\% respectively.

\begin{figure}[!ht]
\centering
\includegraphics[width=\textwidth]{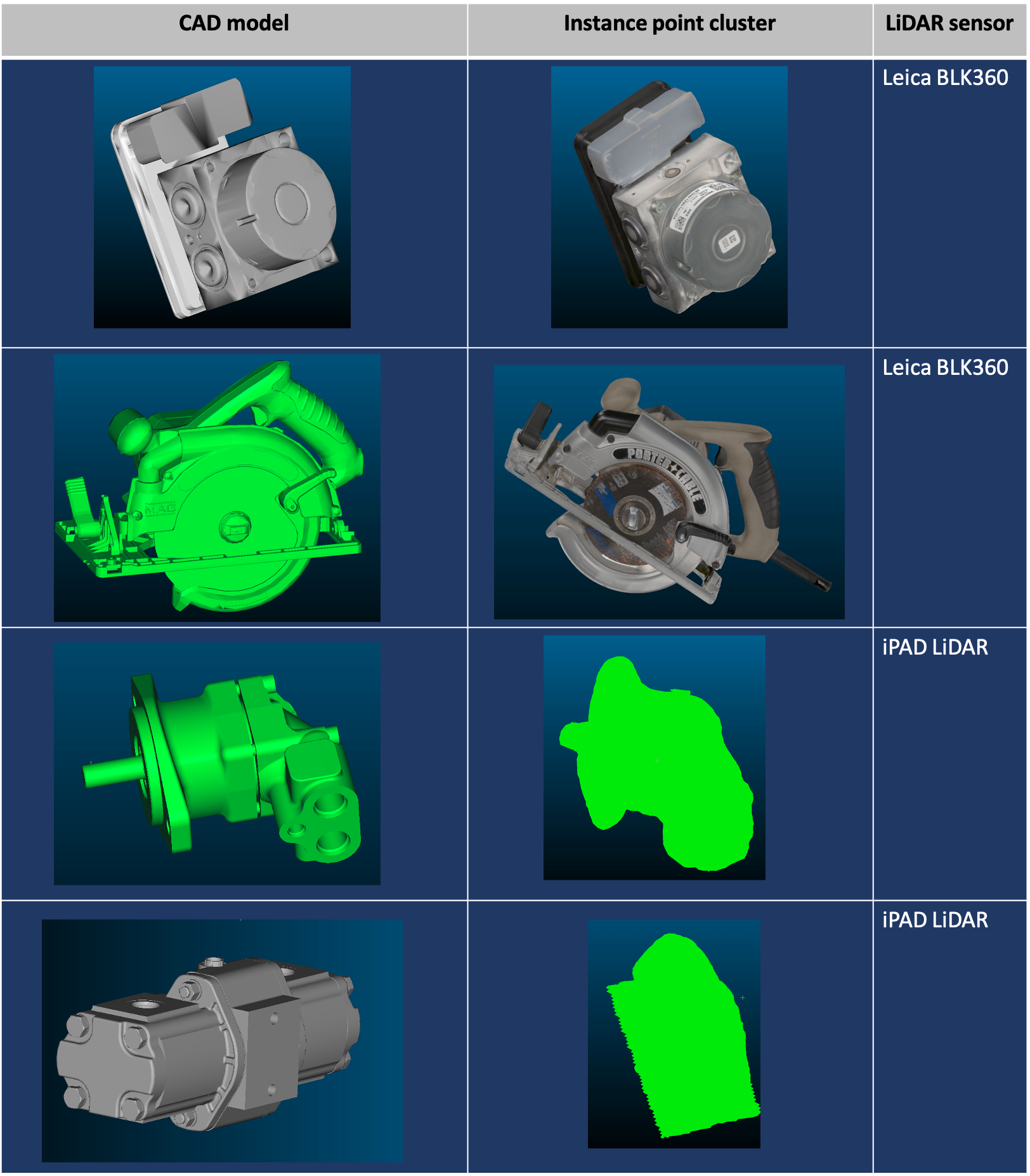}
\caption{CAD models and scanned instance point clusters.}
\label{fig:scannedobjects}
\end{figure}

\begin{figure}[!ht]
\centering
\includegraphics[width=\textwidth]{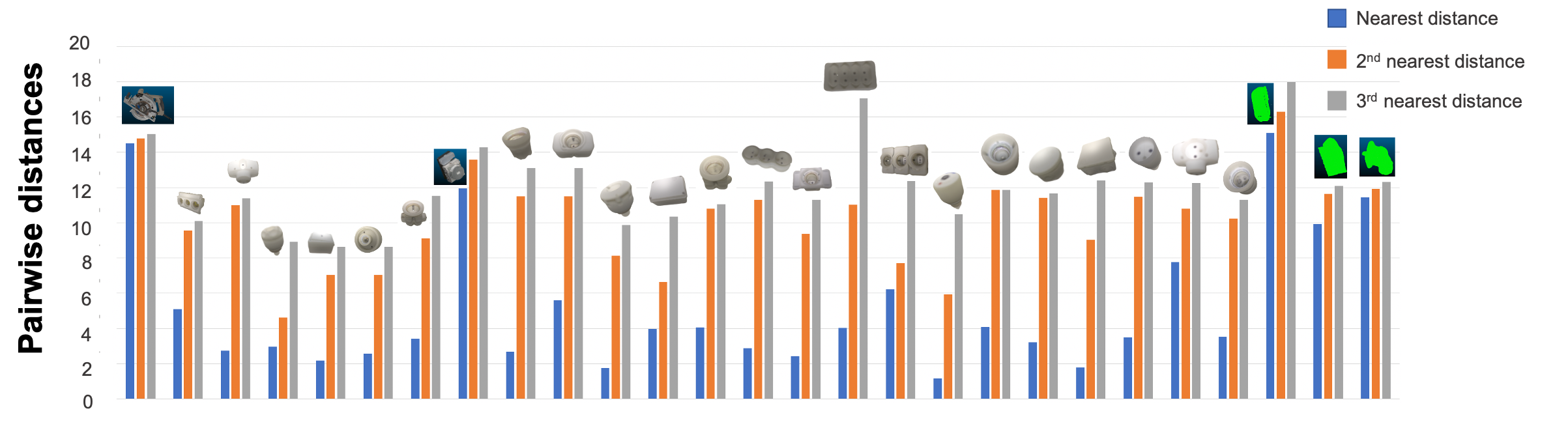}
\caption{Distribution of pairwise distances for each T-LESS mechanical object and laser scanned objects.}
\label{fig:pairwise}
\end{figure}

\begin{figure}[!ht]
\centering
\includegraphics[width=\textwidth]{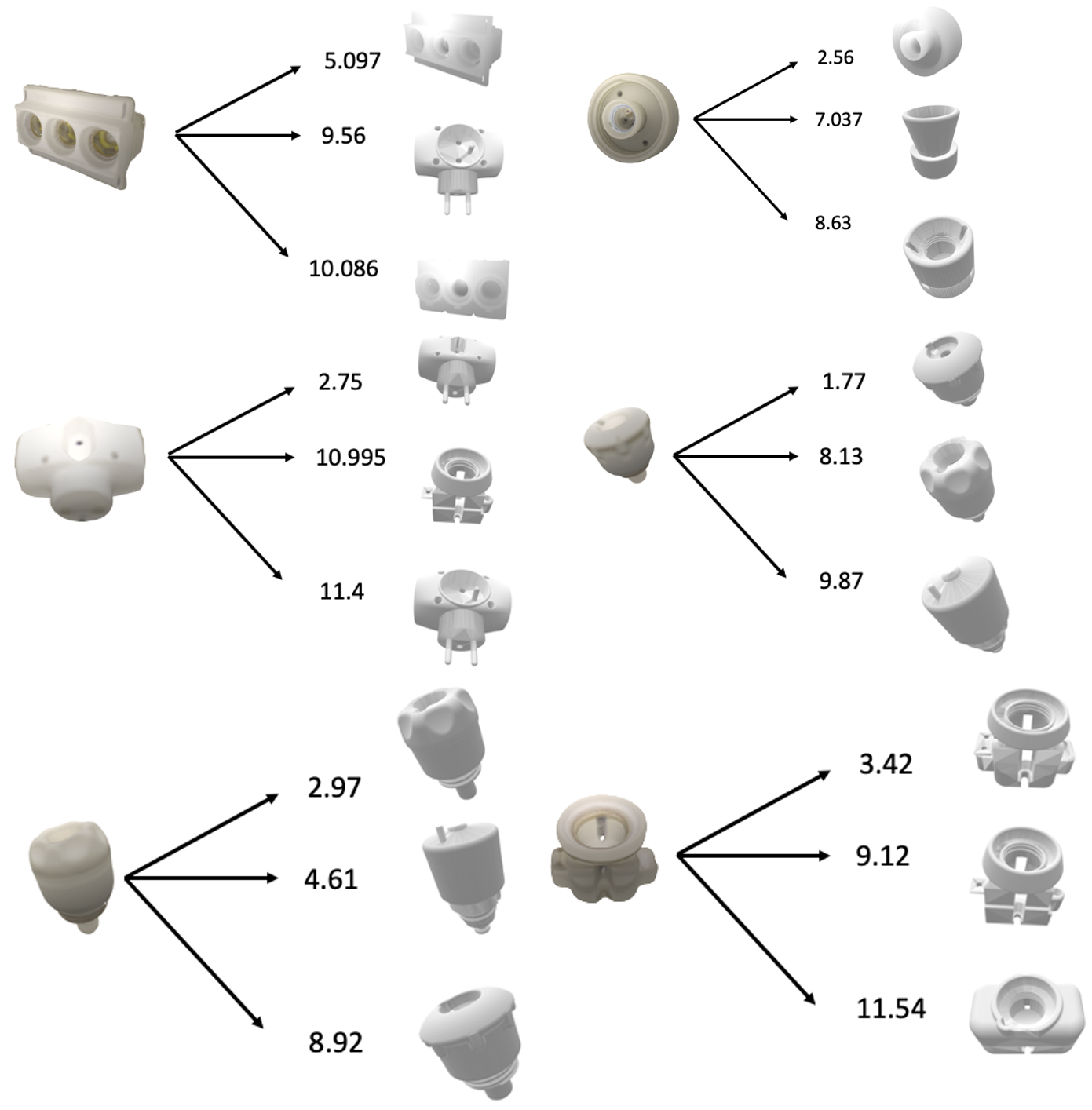}
\caption{Examples of T-LESS reconstructed objects and their retrieved CAD models with illustrated pairwise Euclidean distances.}
\label{fig:exampledistances}
\end{figure}

{\bf Sensitivity to laser scanned instance segments and partial scans}. We scanned four industrial objects, two of them with a Leica BLK360 and two with an iPAD LiDAR (shown in Figure \ref{fig:scannedobjects}) in order to quantify the impact of the laser scan quality on the success of the shape retrieval method. We also plot the distribution of the pairwise distances (Euclidean feature vector distance between the CAD and reconstructed model). Pairwise distances represent the distance between two feature vectors; the feature vector of the laser scanned object and the feature vector of a CAD model are dimensionless. Figure \ref{fig:pairwise} shows that pairwise distances for T-LESS objects are typically less than 12 (reconstructed objects), whereas the industrial objects scanned by LiDAR sensors, have larger distances (more than 10 on average). The nearest retrieved shape for the saw object is at a distance of 14 (one of the largest distances in Figure \ref{fig:pairwise}). Further research needs to be conducted to identify the impact and quality of scanned objects to shape retrieval given the sensitivity of feature embeddings on input data quality. Recently, \cite{Kim2002DimensionalScans} identified 18 piping components at 79.15\% accuracy, however more data was needed to generalize the approach. One of the main goals of this work is to validate the approach on ``perfect'' data before validating it with imperfect, occluded and partial scans. Figure \ref{fig:exampledistances} illustrates six T-LESS reconstructed components and their retrieved CAD models. The pairwise distances for the nearest retrieval/match are less than 10. 

\begin{figure}[!ht]
\centering
\includegraphics[width=0.9\textwidth]{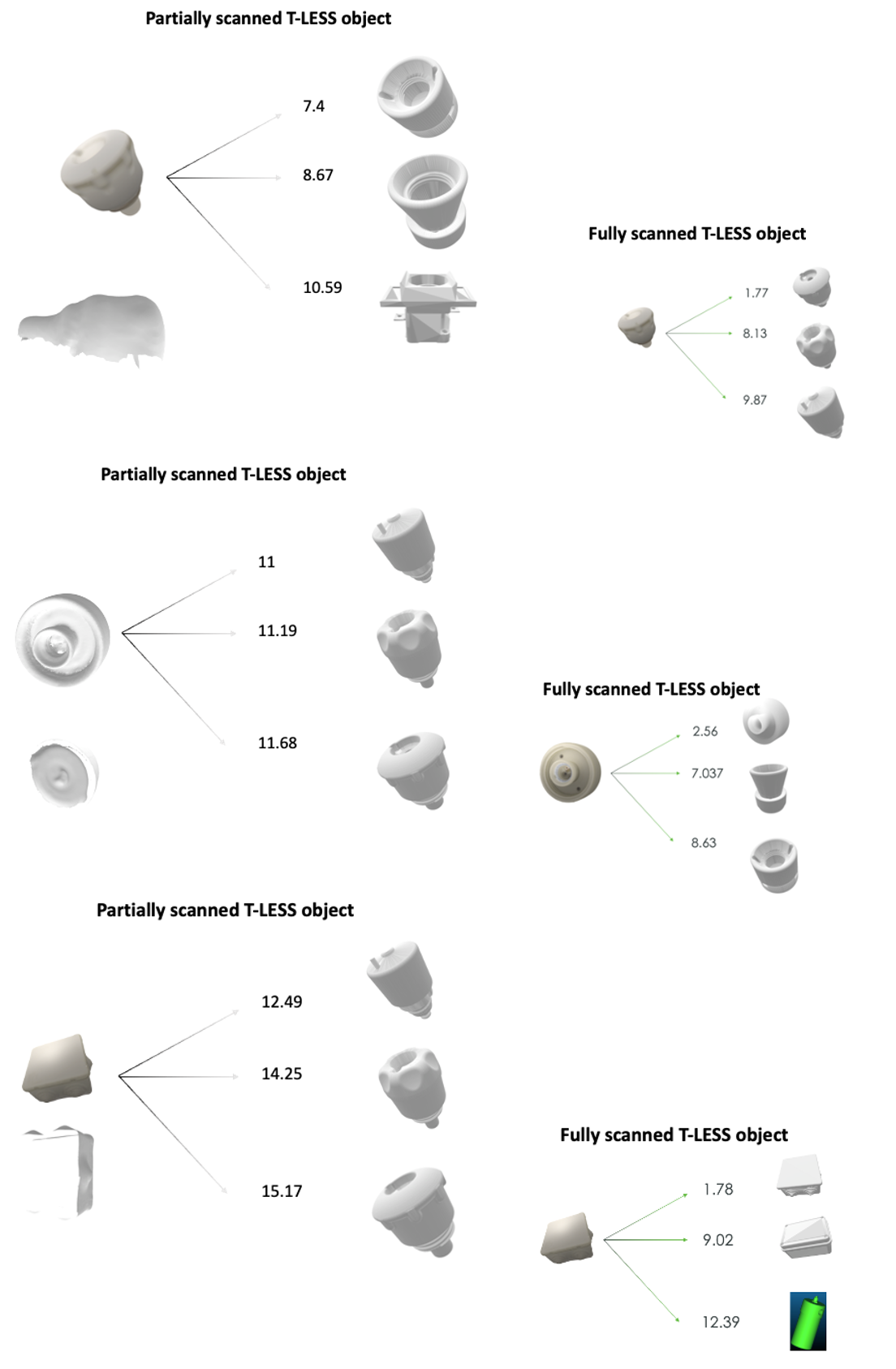}
\caption{Examples of T-LESS partially scanned objects and their retrieved CAD models with illustrated pairwise Euclidean distances.}
\label{fig:partiallyscan}
\end{figure}

We also conducted a sensitivity analysis on the effect of partially scanned objects to the retrieval rate. Figure \ref{fig:partiallyscan} illustrates three T-LESS objects and their respective retrieved CAD models. The pairwise distances of the partially scanned objects for the closest retrieval/match are increased by approximately 300\%-600\%. The difference in pairwise distances is not so significant for the 2nd and 3rd closest retrieval. It is also noteworthy to mention that the correct CAD models could not be retrieved for the partially scanned objects. Further investigation of the partial scanning effects is out of scope of this research.

\begin{figure}[!ht]
\centering
\includegraphics[width=\textwidth]{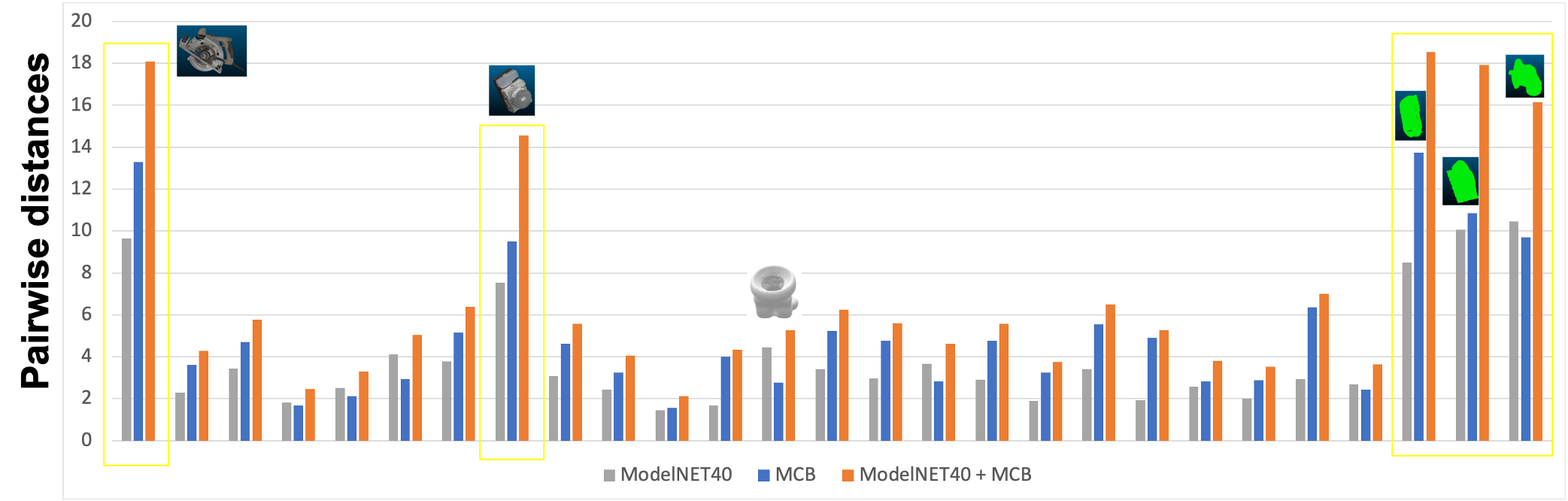}
\caption{Distribution of pairwise distances for each mechanical object and laser scanned objects - Sensitivity analysis on pre-trained network.}
\label{fig:sensitivity}
\end{figure}

{\bf Sensitivity to pre-trained classification network}. We investigate the impact of the selection of multiple pre-trained networks on the pairwise Euclidean distances between CAD models and laser scans. Figure \ref{fig:sensitivity} shows the pairwise distance distributions when training with three distinctive pre-trained networks: (a) ModelNET40, (b) MCB (B) and (c) ModelNET40 and MCB (B) jointly. We observe that the ModelNET40 network has greater impact on the LiDAR scanned objects rather than the T-LESS reconstructed objects. The results indicate that adding more training data does not necessarily improve the retrieval results (e.g. the combined ModelNET40 and MCB (B) pre-trained network has larger pairwise distances compared to the other two networks). The combined ModelNET40 and MCB (B) datasets have 65 classes in total. 

\begin{figure}[!ht]
\centering
\includegraphics[width=\textwidth]{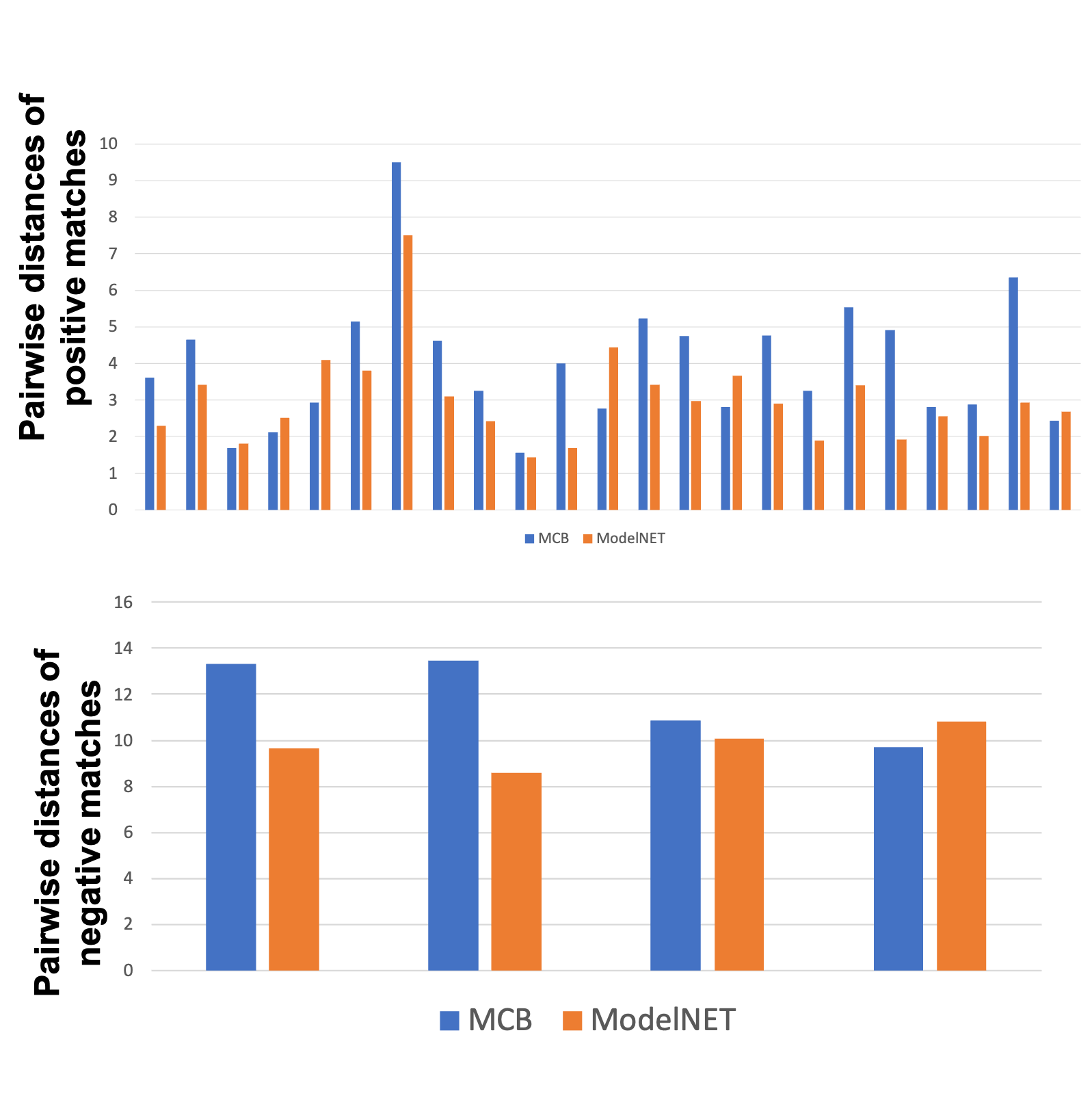}
\caption{Distribution of pairwise distances for positive (top) and negative (bottom) matches.}
\label{fig:positivenegative}
\end{figure}

The previous plots only identified the pairwise distances. However, it is important to investigate the correlation between pairwise distances and positive or negative shape retrievals/matches. We define a positive match as a correct CAD model retrieval for the laser scanned object of interest. Figure \ref{fig:positivenegative} shows that positive matches do not exceed a pairwise distance of 10, whereas negative matches are in the range of 10-14. In other words, close (in terms of Euclidean distance) feature embeddings indicate correct matches.

\section{Conclusions}
\label{conclusions}

This paper presents an automated shape retrieval methodology that has the potential to extend the benchmarked CLOI framework for generating gDTs of existing industrial facilities from point cloud data. This work focuses on the identification of the CAD shape which is closest (in the feature space) to segmented instance point clusters. In the following paragraphs, we present the contributions ({\bf Con}) and limitations ({\bf Lim}) of the proposed shape retrieval method in detail.

{\bf Con 1} This is the first methodology of its kind to achieve significantly high performance (85.2\% overall accuracy) compared to current state-of-the-art research. It, therefore, provides a solid foundation for future work in generating gDTs of industrial facilities. {\bf Con 2} This research moves forward the state of automated shape retrieval from ``ideally'' segmented instance point clusters as well as promotes the value of adding ``intelligence'' to the point cloud data. The interpretation of the results indicate that the performance of the method is sensitive to partial scans and quality of input laser scanned data.  {\bf Con 3} The simple yet effective joint network leverages for the first time geometric and visual features in classifying instances of industrial objects and has the potential to be extended to other tasks (such as class and instance segmentation). {\bf Con 4} This paper presents a comprehensive overview of geometric digital twinning SOTA research and practice as well as benchmarks a gDT framework for further experimentation.

{\bf Lim 1} The shape retrieval method has been primarily tested on pre-segmented instance point clusters. Therefore, there is a need to quantify the success of the method jointly with our previous research on the CLOI framework \cite{Agapaki2020}. {\bf Lim 2} As demonstrated in Figure \ref{fig:sensitivity}, more training data is not always beneficial, so a careful experimental set-up should be conducted to avoid negative impacts on retrieval performance. {\bf Lim 3} Image augmentation for the image classification network to more than four images needs to performed and validated with the current study. In these efforts, the minimum data requirements should be determined for effective shape retrieval. 

Applicability of this research extends to part identification in industrial catalogs, better inventory management and improved gDT workflows. 

There are several gaps in knowledge around industrial gDT in research that follows based on the findings of this work and would benefit from further research, to extend and further enhance the developed framework. Direct future research includes: (a) an improved joint network framework with additional SOTA classification networks and (b) shape retrieval as part of the CLOI framework where the automatically segmented instances will be the input of the proposed method.

\subsection{DATA AVAILABILITY}

Some or all data, models, or code used during the study were provided by a third party. Direct requests for these materials may be made to the provider as indicated in the Acknowledgements.

\subsection{ACKNOWLEDGEMENTS}

We thank our colleague Graham Miatt, who has provided insight, expertise and data that greatly assisted this research. We also express our gratitude to Bob Flint from BP International Centre for Business and Technology (ICBT), who provided data for evaluation. The research leading to these results has received funding from the Engineering and Physical Sciences Research Council (EPSRC) and the US National Academy of Engineering (NAE). AVEVA Group Plc. and BP International Centre for Business and Technology (ICBT) partially sponsor this research under grant agreements RG83104 and RG90532 respectively. We gratefully acknowledge the collaboration of all academic and industrial project partners. Any opinions, findings and conclusions or recommendations expressed in this material are those of the authors and do not necessarily reflect the views of the institutes mentioned above.

\begin{table}[!ht]
\centering
\caption{Dataset of mechanical components (MCB-B) and their specifications}
    \begin{tabular}{ l l } 
    \hline\hline
    {\bf Class} & {\bf Number of CAD models} \\
    \hline
    \ Bearings & 1,117 \\ 
    \ Bushes & 592 \\
    \ Castors & 1,109 \\
    \ Clamps & 157 \\
    \ Discs & 109 \\
    \ Fittings & 1,756 \\
    \ Flanges & 398 \\
    \ Fork joints & 47 \\
    \ Gears & 515 \\
    \ Handles & 1,751 \\
    \ Hinges & 61 \\
    \ Hooks & 122 \\
    \ Motors & 746 \\
    \ Nuts & 1,069 \\
    \ Pins & 2,659 \\
    \ Plates & 366 \\
    \ Pullies & 312 \\
    \ Rings & 551 \\
    \ Rivets & 51 \\
    \ Rotors & 470 \\
    \ Screws & 3,661 \\
    \ Springs & 348 \\
    \ Studs & 352 \\
    \ Switches & 177 \\
    \ Washers & 880 \\
    \hline
    \hline
    \end{tabular}
\label{table:benchmark}
\end{table}

\begin{table}[!ht]
\centering
\caption{Performance of deep learning frameworks on MCB(B) dataset}
\large
\begin{tabular}{ l l l l} 
\hline\hline
{\bf Method} & {\bf \makecell{Precision \\ (\%)}} & {\bf \makecell{Recall \\ (\%)}} & {\bf \makecell{Accuracy \\ (\%)}} \\
\hline
\ {\bf PointNET \cite{Qi2017PointNET:Segmentation}} & 66.9 & 61.1 & 63 \\
\ {\bf MobileNETv2 \cite{sandler2018mobilenetv2}} & 60.6 & 54.1 & 64 \\
\ {\bf Joint network (proposed)} & 74.6 & 71.7 & 85.3 \\
\hline\hline
\end{tabular}
\label{table:classificationnetworks}
\end{table}

\begin{table}[!ht]
\centering
\caption{Recognition rates on TLESS dataset}
\large
\begin{tabular}{ l l l} 
\hline\hline
{\bf \makecell{Top 1 accuracy \\ (\%)}} & {\bf \makecell{Top 3 accuracy \\ (\%)}} & {\bf \makecell{Top 5 accuracy \\ (\%)}} \\
\hline
\ 85.2 & 88.9 & 88.9 \\
\hline\hline
\end{tabular}
\label{table:classificationnetworks}
\end{table}

\bibliography{references}

\begin{thebibliography}{}

\bibitem[\protect\citeauthoryear{}{Abdel-Malek
  et~al.\@}{2001}]{Abdel-Malek2001OnSurfaces}
Abdel-Malek, K., Yang, J., and Blackmore, D. (2001).
\newblock ``{On swept volume formulations: Implicit surfaces}.''\ {\em CAD
  Computer Aided Design}.

\bibitem[\protect\citeauthoryear{}{Abdel-Malek and
  Yeh}{1997}]{Abdel-Malek1997GeometricConditions}
Abdel-Malek, K. and Yeh, H.~J. (1997).
\newblock ``{Geometric representation of the swept volume using Jacobian
  rank-deficiency conditions}.''\ {\em CAD Computer Aided Design}.

\bibitem[\protect\citeauthoryear{}{Agapaki}{2020}]{Agapaki2020}
Agapaki, E. (2020).
\newblock ``{Automated Object Segmentation in Existing Industrial
  Facilities}.''\ Doctoral thesis, University of Cambridge, ,
  $<$https://doi.org/10.17863/CAM.52102$>$.

\bibitem[\protect\citeauthoryear{}{Agapaki and
  Brilakis}{2020a}]{agapaki2020cloi}
Agapaki, E. and Brilakis, I. (2020a).
\newblock ``Cloi-net: Class segmentation of industrial facilities’ point
  cloud datasets.''\ {\em Advanced Engineering Informatics}, 45, 101121.

\bibitem[\protect\citeauthoryear{}{Agapaki and
  Brilakis}{2020b}]{agapaki2020instance}
Agapaki, E. and Brilakis, I. (2020b).
\newblock ``Instance segmentation of industrial point cloud data.

\bibitem[\protect\citeauthoryear{}{Agapaki
  et~al.\@}{2019}]{Agapaki2019CLOI:Facilities}
Agapaki, E., Glyn-Davies, A., Mandoki, S., and Brilakis, I. (2019).
\newblock ``{CLOI: A Shape Classification Benchmark Dataset for Industrial
  Facilities}.''\ {\em 2019 ASCE International Conference on Computing in Civil
  Engineering}.

\bibitem[\protect\citeauthoryear{}{Agapaki
  et~al.\@}{2018}]{Agapaki2018PrioritizingFacilities}
Agapaki, E., Miatt, G., and Brilakis, I. (2018).
\newblock ``{Prioritizing object types for modelling existing industrial
  facilities}.''\ {\em Automation in Construction}.

\bibitem[\protect\citeauthoryear{}{Agapaki and
  Nahangi}{2020}]{Agapaki2020ChapterGeneration}
Agapaki, E. and Nahangi, M. (2020).
\newblock ``{Chapter 3 - Scene understanding and model generation}.''\ {\em
  Infrastructure Computer Vision}, I. Brilakis and C. Haas, eds., Elsevier, 1
  edition, Chapter~3.

\bibitem[\protect\citeauthoryear{}{Alexa
  et~al.\@}{2001}]{Alexa2001PointSurfaces}
Alexa, M., Behr, J., Cohen-Or, D., Fleishman, S., Levin, D., and Silva, C.~T.
  (2001).
\newblock ``{Point set surfaces}.''\ {\em Proceedings of the IEEE Visualization
  Conference}.

\bibitem[\protect\citeauthoryear{}{Amenta et~al.\@}{2001}]{Amenta2001TheCrust}
Amenta, N., Choi, S., and Kolluri, R.~K. (2001).
\newblock ``{The power crust}.''\ {\em Proceedings of the Symposium on Solid
  Modeling and Applications}.

\bibitem[\protect\citeauthoryear{}{Barbosa
  et~al.\@}{2017}]{Barbosa2017ReinventingProductivity}
Barbosa, F., Woetzel, J., Mischke, J., Joao~Ribeirinho, M., Sridhar, M.,
  Parsons, M., and Brown, S. (2017).
\newblock ``{Reinventing Construction: A Route to Higher Productivity},
  $<$https://www.mckinsey.com/~/media/McKinsey/Industries/Capital Projects and
  Infrastructure/Our Insights/Reinventing construction through a productivity
  revolution/MGI-Reinventing-construction-A-route-to-higher-productivity-Full-report.ashx$>$.

\bibitem[\protect\citeauthoryear{}{Barnhill}{1977}]{Barnhill1977RepresentationSurfaces}
Barnhill, R.~E. (1977).
\newblock ``{Representation and Approximation of Surfaces}.''\ {\em
  Mathematical Software}.

\bibitem[\protect\citeauthoryear{}{Besl and McKay}{1992}]{Besl1992AShapes}
Besl, P.~J. and McKay, N.~D. (1992).
\newblock ``{A method for registration of 3-D shapes}.''\ {\em IEEE
  Transactions on Pattern Analysis and Machine Intelligence}, 14(2), 239--256.

\bibitem[\protect\citeauthoryear{}{Blackmore
  et~al.\@}{1999}]{Blackmore1999TrimmingVolumes}
Blackmore, D., Samulyak, R., and Leu, M.~C. (1999).
\newblock ``{Trimming swept volumes}.''\ {\em CAD Computer Aided Design}.

\bibitem[\protect\citeauthoryear{}{Bosch{\'{e}}
  et~al.\@}{2012}]{Bosche2012MarkerlessVisualization}
Bosch{\'{e}}, F., Tingdahl, D., Carozza, L., and Van~Gool, L. (2012).
\newblock ``{Markerless vision-based Augmented Reality for enhanced project
  visualization}.''\ {\em 2012 Proceedings of the 29th International Symposium
  of Automation and Robotics in Construction, ISARC 2012}.

\bibitem[\protect\citeauthoryear{}{Boulch
  et~al.\@}{2017}]{boulch2017unstructured}
Boulch, A., Le~Saux, B., and Audebert, N. (2017).
\newblock ``Unstructured point cloud semantic labeling using deep segmentation
  networks..''\ {\em 3DOR@ Eurographics}, 3.

\bibitem[\protect\citeauthoryear{}{Chen and
  Hung}{1999}]{Chen1999RANSAC-BasedImages}
Chen, C.~S. and Hung, Y.~P. (1999).
\newblock ``{RANSAC-Based DARCES: A new approach to fast automatic registration
  of partially overlapping range images}.''\ {\em IEEE Transactions on Pattern
  Analysis and Machine Intelligence}.

\bibitem[\protect\citeauthoryear{}{Chen et~al.\@}{2018}]{chen2018performance}
Chen, J., Fang, Y., and Cho, Y.~K. (2018).
\newblock ``Performance evaluation of 3d descriptors for object recognition in
  construction applications.''\ {\em Automation in Construction}, 86, 44--52.

\bibitem[\protect\citeauthoryear{}{Chen
  et~al.\@}{2017}]{Chen2017Geometry-basedManagement}
Chen, J., Zhang, C., and Tang, P. (2017).
\newblock ``{Geometry-based optimized point cloud compression methodology for
  construction and infrastructure management}.''\ {\em Congress on Computing in
  Civil Engineering, Proceedings}.

\bibitem[\protect\citeauthoryear{}{Chow and
  Rad}{2002}]{Chow2002On-lineAlgorithms}
Chow, K.~M. and Rad, A.~B. (2002).
\newblock ``{On-line fuzzy identification using genetic algorithms}.''\ {\em
  Fuzzy Sets and Systems}.

\bibitem[\protect\citeauthoryear{}{Choy et~al.\@}{2019}]{Choy20194DNetworks}
Choy, C., Gwak, J., and Savarese, S. (2019).
\newblock ``{4D Spatio-Temporal ConvNets: Minkowski Convolutional Neural
  Networks}.''\ {\em Computer Vision and Pattern Recognition (CVPR)}.

\bibitem[\protect\citeauthoryear{}{Chung
  et~al.\@}{1998}]{Chung1998RegistrationTechnique}
Chung, D.~H., Yun, I.~D., and Lee, S.~U. (1998).
\newblock ``{Registration of multiple-range views using the reverse-calibration
  technique}.''\ {\em Pattern Recognition}.

\bibitem[\protect\citeauthoryear{}{{Cintoo}}{2022}]{Cintoo}
{Cintoo} (2022).
\newblock ``Cintoo cloud: Bridging the gap between reality capture and digital
  twins.

\bibitem[\protect\citeauthoryear{}{{ClearEdge}}{2019}]{ClearEdge2019PlantCapabilities}
{ClearEdge} (2019).
\newblock ``{Plant Modeling Capabilities},
  $<$https://www.clearedge3d.com/products/edgewise-plant/$>$.

\bibitem[\protect\citeauthoryear{}{Davis
  et~al.\@}{2007}]{10.1145/1273496.1273523}
Davis, J.~V., Kulis, B., Jain, P., Sra, S., and Dhillon, I.~S. (2007).
\newblock ``Information-theoretic metric learning.''\ {\em Proceedings of the
  24th International Conference on Machine Learning}, ICML '07, New York, NY,
  USA, Association for Computing Machinery,  209–216,
  $<$https://doi.org/10.1145/1273496.1273523$>$.

\bibitem[\protect\citeauthoryear{}{{Dawson-Haggerty et al.}}{}]{trimesh}
{Dawson-Haggerty et al.}
\newblock ``trimesh, $<$https://trimsh.org/$>$.

\bibitem[\protect\citeauthoryear{}{Dey and
  Goswami}{2003}]{Dey2003TightReconstructor}
Dey, T.~K. and Goswami, S. (2003).
\newblock ``{Tight Cocone: A Water-tight Surface Reconstructor}.''\ {\em
  Journal of Computing and Information Science in Engineering}.

\bibitem[\protect\citeauthoryear{}{Dey and Sun}{2005}]{Dey2005AnGuarantees}
Dey, T.~K. and Sun, J. (2005).
\newblock ``{An Adaptive MLS Surface for Reconstruction with Guarantees}.''\
  {\em Eurographics Symposium on Geometry Processing}.

\bibitem[\protect\citeauthoryear{}{Eck
  et~al.\@}{1995}]{Eck1995MultiresolutionMeshes}
Eck, M., DeRose, T., Duchamp, T., Hoppe, H., Lounsbery, M., and Stuetzle, W.
  (1995).
\newblock ``{Multiresolution analysis of arbitrary meshes}.''\ {\em Proceedings
  of the ACM SIGGRAPH Conference on Computer Graphics}.

\bibitem[\protect\citeauthoryear{}{Edelsbrunner and
  Miicke}{1992}]{Edelsbrunner1992Three-dimensionalShapes}
Edelsbrunner, H. and Miicke, E.~P. (1992).
\newblock ``{Three-dimensional alpha shapes}.''\ {\em Proceedings of the 1992
  Workshop on Volume Visualization, VVS 1992}.

\bibitem[\protect\citeauthoryear{}{Fan et~al.\@}{2021}]{fan2021scf}
Fan, S., Dong, Q., Zhu, F., Lv, Y., Ye, P., and Wang, F.-Y. (2021).
\newblock ``Scf-net: Learning spatial contextual features for large-scale point
  cloud segmentation.''\ {\em Proceedings of the IEEE/CVF Conference on
  Computer Vision and Pattern Recognition},  14504--14513.

\bibitem[\protect\citeauthoryear{}{Fleishman
  et~al.\@}{2005}]{Fleishman2005RobustFeatures}
Fleishman, S., Cohen-Or, D., and Silva, C.~T. (2005).
\newblock ``{Robust moving least-squares fitting with sharp features}.''\ {\em
  ACM Transactions on Graphics}.

\bibitem[\protect\citeauthoryear{}{Floater}{1997}]{Floater1997ParametrizationTriangulations}
Floater, M.~S. (1997).
\newblock ``{Parametrization and smooth approximation of surface
  triangulations}.''\ {\em Computer Aided Geometric Design}.

\bibitem[\protect\citeauthoryear{}{Franke and
  Nielson}{1991}]{Franke1991ScatteredSurvey}
Franke, R. and Nielson, G.~M. (1991).
\newblock ``{Scattered Data Interpolation and Applications: A Tutorial and
  Survey}.

\bibitem[\protect\citeauthoryear{}{Fumarola and
  Poelman}{2011}]{Fumarola2011GeneratingApproaches}
Fumarola, M. and Poelman, R. (2011).
\newblock ``{Generating virtual environments of real world facilities:
  Discussing four different approaches}.''\ {\em Automation in Construction},
  Vol.~20,  263--269.

\bibitem[\protect\citeauthoryear{}{Gerardo-Castro
  et~al.\@}{2015}]{Gerardo-Castro2015Laser-radarSurfaces}
Gerardo-Castro, M.~P., Peynot, T., and Ramos, F. (2015).
\newblock ``{Laser-radar data fusion with Gaussian process implicit
  surfaces}.''\ {\em Springer Tracts in Advanced Robotics}.

\bibitem[\protect\citeauthoryear{}{Gerardo-Castro
  et~al.\@}{2014}]{Gerardo-Castro2014RobustSurfaces}
Gerardo-Castro, M.~P., Peynot, T., Ramos, F., and Fitch, R. (2014).
\newblock ``{Robust multiple-sensing-modality data fusion using Gaussian
  Process Implicit Surfaces}.''\ {\em FUSION 2014 - 17th International
  Conference on Information Fusion}.

\bibitem[\protect\citeauthoryear{}{Gerbert
  et~al.\@}{2016}]{Gerbert2016DigitalConstruction}
Gerbert, P., Castagnino, S., Rothballer, C., Renz, A., and Filitz, R. (2016).
\newblock ``{Digital in Engineering and Construction},
  $<$http://futureofconstruction.org/content/uploads/2016/09/BCG-Digital-in-Engineering-and-Construction-Mar-2016.pdf$>$.

\bibitem[\protect\citeauthoryear{}{{GI Hub}}{2017}]{GIHub2017}
{GI Hub} (2017).
\newblock ``Global infrastructure outlook, global infrastructure hub.

\bibitem[\protect\citeauthoryear{}{Graham et~al.\@}{2018}]{graham20183d}
Graham, B., Engelcke, M., and Van Der~Maaten, L. (2018).
\newblock ``3d semantic segmentation with submanifold sparse convolutional
  networks.''\ {\em Proceedings of the IEEE conference on computer vision and
  pattern recognition},  9224--9232.

\bibitem[\protect\citeauthoryear{}{Greenspan and
  Godin}{2001}]{Greenspan2001AICP}
Greenspan, M. and Godin, G. (2001).
\newblock ``{A nearest neighbor method for efficient ICP}.''\ {\em Proceedings
  of International Conference on 3-D Digital Imaging and Modeling, 3DIM}.

\bibitem[\protect\citeauthoryear{}{Greiner and
  Hormann}{1997}]{Greiner1997InterpolatingB-splines}
Greiner, G. and Hormann, K. (1997).
\newblock ``{Interpolating and approximating scattered 3D-data with
  hierarchical tensor product B-splines}.''\ {\em Surface Fitting and
  Multiresolution Methods}.

\bibitem[\protect\citeauthoryear{}{Guennebaud and
  Gross}{2007}]{Guennebaud2007AlgebraicSurfaces}
Guennebaud, G. and Gross, M. (2007).
\newblock ``{Algebraic point set surfaces}.''\ {\em Proceedings of the ACM
  SIGGRAPH Conference on Computer Graphics}.

\bibitem[\protect\citeauthoryear{}{Gu{\'e}rin and
  Boots}{2018}]{guerin2018improving}
Gu{\'e}rin, J. and Boots, B. (2018).
\newblock ``Improving image clustering with multiple pretrained cnn feature
  extractors.''\ {\em arXiv preprint arXiv:1807.07760}.

\bibitem[\protect\citeauthoryear{}{Hardy}{1971}]{Hardy1971MultiquadricSurfaces}
Hardy, R.~L. (1971).
\newblock ``{Multiquadric equations of topography and other irregular
  surfaces}.''\ {\em Journal of Geophysical Research}.

\bibitem[\protect\citeauthoryear{}{Hodan et~al.\@}{2017}]{hodan2017t}
Hodan, T., Haluza, P., Obdr{\v{z}}{\'a}lek, {\v{S}}., Matas, J., Lourakis, M.,
  and Zabulis, X. (2017).
\newblock ``T-less: An rgb-d dataset for 6d pose estimation of texture-less
  objects.''\ {\em 2017 IEEE Winter Conference on Applications of Computer
  Vision (WACV)}, IEEE,  880--888.

\bibitem[\protect\citeauthoryear{}{Hu et~al.\@}{2020}]{hu2020randla}
Hu, Q., Yang, B., Xie, L., Rosa, S., Guo, Y., Wang, Z., Trigoni, N., and
  Markham, A. (2020).
\newblock ``Randla-net: Efficient semantic segmentation of large-scale point
  clouds.''\ {\em Proceedings of the IEEE/CVF Conference on Computer Vision and
  Pattern Recognition},  11108--11117.

\bibitem[\protect\citeauthoryear{}{Hullo
  et~al.\@}{2015}]{Hullo2015Multi-SensorArchitectures}
Hullo, J.-F., Thibault, G., Boucheny, C., Dory, F., and Mas, A. (2015).
\newblock ``{Multi-Sensor As-Built Models of Complex Industrial
  Architectures}.''\ {\em Remote Sensing}, 7(12), 16339--16362.

\bibitem[\protect\citeauthoryear{}{{IET}}{2019}]{IET2019DigitalEnvironment}
{IET} (2019).
\newblock ``{Digital Twins for the built environment}.''\ {\em Report no.}, The
  Institution of Engineering and Technology,
  $<$https://www.theiet.org/media/4719/digital-twins-for-the-built-environment.pdf$>$.

\bibitem[\protect\citeauthoryear{}{Joy}{1992}]{Joy1992VisualizationSolids}
Joy, K.~I. (1992).
\newblock ``{Visualization of Swept Hyperpatch Solids}.''\ {\em Visual
  Computing}.

\bibitem[\protect\citeauthoryear{}{Joy and
  Duchaineau}{1999}]{Joy1999BoundarySolids}
Joy, K.~I. and Duchaineau, M.~A. (1999).
\newblock ``{Boundary determination for trivariate solids}.''\ {\em Proceedings
  - 7th Pacific Conference on Computer Graphics and Applications, Pacific
  Graphics 1999}.

\bibitem[\protect\citeauthoryear{}{Kim et~al.\@}{2013}]{Kim2013FullyMonitoring}
Kim, C., Kim, C., and Son, H. (2013).
\newblock ``{Fully automated registration of 3D data to a 3D CAD model for
  project progress monitoring}.''\ {\em Automation in Construction}, 35,
  587--594.

\bibitem[\protect\citeauthoryear{}{Kim
  et~al.\@}{2002}]{Kim2002DimensionalScans}
Kim, H., Haas, C.~T., Rauch, A.~F., and Browne, C. (2002).
\newblock ``{Dimensional Ratios for Stone Aggregates from Three-Dimensional
  Laser Scans}.''\ {\em Journal of Computing in Civil Engineering}.

\bibitem[\protect\citeauthoryear{}{Kim et~al.\@}{2020a}]{kim2020deep}
Kim, H., Yeo, C., Lee, I.~D., and Mun, D. (2020a).
\newblock ``Deep-learning-based retrieval of piping component catalogs for
  plant 3d cad model reconstruction.''\ {\em Computers in Industry}, 123,
  103320.

\bibitem[\protect\citeauthoryear{}{Kim et~al.\@}{2020b}]{sangpil2020large}
Kim, S., Chi, H.-g., Hu, X., Huang, Q., and Ramani, K. (2020b).
\newblock ``A large-scale annotated mechanical components benchmark for
  classification and retrieval tasks with deep neural networks.''\ {\em
  Proceedings of 16th European Conference on Computer Vision (ECCV)}.

\bibitem[\protect\citeauthoryear{}{Komori and Hotta}{2019}]{komori2019ab}
Komori, J. and Hotta, K. (2019).
\newblock ``Ab-pointnet for 3d point cloud recognition.''\ {\em 2019 Digital
  Image Computing: Techniques and Applications (DICTA)}, IEEE,  1--6.

\bibitem[\protect\citeauthoryear{}{Lawin et~al.\@}{2017}]{lawin2017deep}
Lawin, F.~J., Danelljan, M., Tosteberg, P., Bhat, G., Khan, F.~S., and
  Felsberg, M. (2017).
\newblock ``Deep projective 3d semantic segmentation.''\ {\em International
  Conference on Computer Analysis of Images and Patterns}, Springer,  95--107.

\bibitem[\protect\citeauthoryear{}{Lee
  et~al.\@}{2013}]{Lee2013Skeleton-basedData}
Lee, J., Son, H., Kim, C., and Kim, C. (2013).
\newblock ``{Skeleton-based 3D reconstruction of as-built pipelines from
  laser-scan data}.''\ {\em Automation in Construction}, 35, 199--207.

\bibitem[\protect\citeauthoryear{}{Li et~al.\@}{2019a}]{Li2019SupervisedClouds}
Li, L., Sung, M., Dubrovina, A., Yi, L., and Guibas, L. (2019a).
\newblock ``{Supervised Fitting of Geometric Primitives to 3D Point Clouds}.''\
  {\em CVPR}.

\bibitem[\protect\citeauthoryear{}{Li et~al.\@}{2019b}]{li2019tgnet}
Li, Y., Ma, L., Zhong, Z., Cao, D., and Li, J. (2019b).
\newblock ``Tgnet: Geometric graph cnn on 3-d point cloud segmentation.''\ {\em
  IEEE Transactions on Geoscience and Remote Sensing}, 58(5), 3588--3600.

\bibitem[\protect\citeauthoryear{}{Li et~al.\@}{2011}]{Li2011GlobFitRelations}
Li, Y., Wu, X., Chrysathou, Y., Sharf, A., Cohen-Or, D., and Mitra, N.~J.
  (2011).
\newblock ``{GlobFit : Consistently Fitting Primitives by Discovering Global
  Relations}.''\ {\em ACM SIGGRAPH 2011 papers on - SIGGRAPH '11}.

\bibitem[\protect\citeauthoryear{}{Liang et~al.\@}{2021}]{liang2021instance}
Liang, Z., Li, Z., Xu, S., Tan, M., and Jia, K. (2021).
\newblock ``Instance segmentation in 3d scenes using semantic superpoint tree
  networks.''\ {\em Proceedings of the IEEE/CVF International Conference on
  Computer Vision},  2783--2792.

\bibitem[\protect\citeauthoryear{}{Liang
  et~al.\@}{2019}]{liang2019hierarchical}
Liang, Z., Yang, M., Deng, L., Wang, C., and Wang, B. (2019).
\newblock ``Hierarchical depthwise graph convolutional neural network for 3d
  semantic segmentation of point clouds.''\ {\em 2019 International Conference
  on Robotics and Automation (ICRA)}, IEEE,  8152--8158.

\bibitem[\protect\citeauthoryear{}{Lipman
  et~al.\@}{2007}]{Lipman2007Parameterization-freeReconstruction}
Lipman, Y., Cohen-Or, D., Levin, D., and Tal-Ezer, H. (2007).
\newblock ``{Parameterization-free projection for geometry reconstruction}.''\
  {\em ACM Transactions on Graphics}.

\bibitem[\protect\citeauthoryear{}{Liu et~al.\@}{2012}]{liu2012metric}
Liu, E.~Y., Guo, Z., Zhang, X., Jojic, V., and Wang, W. (2012).
\newblock ``Metric learning from relative comparisons by minimizing squared
  residual.''\ {\em 2012 IEEE 12th International Conference on Data Mining},
  IEEE,  978--983.

\bibitem[\protect\citeauthoryear{}{Liu et~al.\@}{2021}]{liu2021beacon}
Liu, T., Cai, Y., Zheng, J., and Thalmann, N.~M. (2021).
\newblock ``Beacon: a boundary embedded attentional convolution network for
  point cloud instance segmentation.''\ {\em The Visual Computer},  1--11.

\bibitem[\protect\citeauthoryear{}{Liu et~al.\@}{2019}]{liu2019relation}
Liu, Y., Fan, B., Xiang, S., and Pan, C. (2019).
\newblock ``Relation-shape convolutional neural network for point cloud
  analysis.''\ {\em Proceedings of the IEEE/CVF Conference on Computer Vision
  and Pattern Recognition},  8895--8904.

\bibitem[\protect\citeauthoryear{}{Liu et~al.\@}{2013}]{Liu2013CylinderPlant}
Liu, Y.~J., Zhang, J.~B., Hou, J.~C., Ren, J.~C., and Tang, W.~Q. (2013).
\newblock ``{Cylinder detection in large-scale point cloud of pipeline
  plant}.''\ {\em IEEE Transactions on Visualization and Computer Graphics},
  19(10), 1700--1707.

\bibitem[\protect\citeauthoryear{}{Liu et~al.\@}{2020}]{liu2020closer}
Liu, Z., Hu, H., Cao, Y., Zhang, Z., and Tong, X. (2020).
\newblock ``A closer look at local aggregation operators in point cloud
  analysis.''\ {\em European Conference on Computer Vision}, Springer,
  326--342.

\bibitem[\protect\citeauthoryear{}{Lu et~al.\@}{2020}]{lu2020pointngcnn}
Lu, Q., Chen, C., Xie, W., and Luo, Y. (2020).
\newblock ``Pointngcnn: Deep convolutional networks on 3d point clouds with
  neighborhood graph filters.''\ {\em Computers \& Graphics}, 86, 42--51.

\bibitem[\protect\citeauthoryear{}{Ma et~al.\@}{2020}]{ma2020semantic}
Ma, J.~W., Czerniawski, T., and Leite, F. (2020).
\newblock ``Semantic segmentation of point clouds of building interiors with
  deep learning: Augmenting training datasets with synthetic bim-based point
  clouds.''\ {\em Automation in Construction}, 113, 103144.

\bibitem[\protect\citeauthoryear{}{Mahalanobis}{1936}]{mahalanobis1936generalized}
Mahalanobis, P.~C. (1936).
\newblock ``On the generalized distance in statistics.''\ National Institute of
  Science of India.

\bibitem[\protect\citeauthoryear{}{MarketsandMarkets}{2019}]{dtmarket}
MarketsandMarkets (2019).
\newblock ``Digital twin market: Industry analysis and market forecast to 2025.

\bibitem[\protect\citeauthoryear{}{Masuda
  et~al.\@}{2015}]{masuda2015reconstruction}
Masuda, H., Niwa, T., Tanaka, I., and Matsuoka, R. (2015).
\newblock ``Reconstruction of polygonal faces from large-scale point-clouds of
  engineering plants.''\ {\em Computer-Aided Design and Applications}, 12(5),
  555--563.

\bibitem[\protect\citeauthoryear{}{{McKinsey}}{2016}]{McKinsey2016}
{McKinsey} (2016).
\newblock ``Bridging global infrastructure gaps.

\bibitem[\protect\citeauthoryear{}{{McKinsey Global
  Institute}}{2021}]{McKinsey2021}
{McKinsey Global Institute} (2021).
\newblock ``Building a more competitive us manufacturing sector.

\bibitem[\protect\citeauthoryear{}{{Microsoft}}{2022}]{Azure2022}
{Microsoft} (2022).
\newblock ``Custom vision: Easily customize your own state-of-the-art computer
  vision models for your unique use case.

\bibitem[\protect\citeauthoryear{}{Nahangi and
  Haas}{2014}]{Nahangi2014AutomatedFabrication}
Nahangi, M. and Haas, C.~T. (2014).
\newblock ``{Automated 3D compliance checking in pipe spool fabrication}.''\
  {\em Advanced Engineering Informatics}, Vol.~28,  360--369.

\bibitem[\protect\citeauthoryear{}{{National Institute of Standards and
  Technology}}{2020}]{NIST2020}
{National Institute of Standards and Technology} (2020).
\newblock ``Economics of manufacturing machinery maintenance; a survey and
  analysis of u.s. costs and benefits.

\bibitem[\protect\citeauthoryear{}{Nielson}{1993}]{Nielson1993ScatteredModeling}
Nielson, G.~M. (1993).
\newblock ``{Scattered Data Modeling}.''\ {\em IEEE Computer Graphics and
  Applications}.

\bibitem[\protect\citeauthoryear{}{{Organization for Economic Co-operation and
  Development}}{2018}]{OECD2018}
{Organization for Economic Co-operation and Development} (2018).
\newblock ``China's belt and road initiative in the global trade, investment
  and finance landscape.

\bibitem[\protect\citeauthoryear{}{{Panda3D}}{}]{panda3d}
{Panda3D}.
\newblock ``Panda3d: The open source framework for 3d rendering and games,
  $<$https://www.panda3d.org/$>$.

\bibitem[\protect\citeauthoryear{}{Peyghambarzadeh
  et~al.\@}{2020}]{peyghambarzadeh2020point}
Peyghambarzadeh, S.~M., Azizmalayeri, F., Khotanlou, H., and Salarpour, A.
  (2020).
\newblock ``Point-planenet: Plane kernel based convolutional neural network for
  point clouds analysis.''\ {\em Digital Signal Processing}, 98, 102633.

\bibitem[\protect\citeauthoryear{}{Pham et~al.\@}{2019}]{Pham2019JSIS3D:Fields}
Pham, Q., Nguyen, D.~T., Hua, B., Roig, G., and Yeung, S. (2019).
\newblock ``{JSIS3D: Joint Semantic-Instance Segmentation of 3D Point Clouds
  with Multi-Task Pointwise Networks and Multi-Value Conditional Random
  Fields}.''\ {\em CVPR}.

\bibitem[\protect\citeauthoryear{}{{PointFuse}}{2022}]{PointFuse}
{PointFuse} (2022).
\newblock ``Pointfuse software.

\bibitem[\protect\citeauthoryear{}{Qi et~al.\@}{2017a}]{Qi2017PointNet++:Space}
Qi, C.~R., Yi, L., Su, H., and Guibas, L.~J. (2017a).
\newblock ``{PointNet++: Deep Hierarchical Feature Learning on Point Sets in a
  Metric Space}.''\ {\em Computer Vision and Pattern Recognition (CVPR)}.

\bibitem[\protect\citeauthoryear{}{Qi
  et~al.\@}{2017b}]{Qi2017PointNET:Segmentation}
Qi, R., Su, H., K., M., and J., G.~L. (2017b).
\newblock ``{PointNET: Deep Learning on Point Sets for 3D Classification and
  Segmentation}.''\ {\em Computer Vision and Pattern Recognition (CVPR)}.

\bibitem[\protect\citeauthoryear{}{Qiu et~al.\@}{2014}]{Qiu2014Pipe-runClouds}
Qiu, R., Zhou, Q.~Y., and Neumann, U. (2014).
\newblock ``{Pipe-run extraction and reconstruction from point clouds}.''\ {\em
  Lecture Notes in Computer Science (including subseries Lecture Notes in
  Artificial Intelligence and Lecture Notes in Bioinformatics)}, Vol. 8691
  LNCS,  17--30.

\bibitem[\protect\citeauthoryear{}{Rabbani and
  Heuvel}{2004}]{Rabbani2004MethodsComparison}
Rabbani, T. and Heuvel, F. V.~D. (2004).
\newblock ``{Methods for fitting CSG models to point clouds and their
  comparison}.''\ {\em ResearchGate}.

\bibitem[\protect\citeauthoryear{}{Rasmussen}{2004}]{Rasmussen2004GaussianLearning}
Rasmussen, C.~E. (2004).
\newblock ``{Gaussian Processes in machine learning}.''\ {\em Lecture Notes in
  Computer Science (including subseries Lecture Notes in Artificial
  Intelligence and Lecture Notes in Bioinformatics)}.

\bibitem[\protect\citeauthoryear{}{{ReScan260}}{2022}]{Rescan}
{ReScan260} (2022).
\newblock ``Rescan: Make decisions about remote locations.

\bibitem[\protect\citeauthoryear{}{Rusinkiewicz and
  Levoy}{2001}]{Rusinkiewicz2001EfficientAlgorithm}
Rusinkiewicz, S. and Levoy, M. (2001).
\newblock ``{Efficient variants of the ICP algorithm}.''\ {\em Proceedings of
  International Conference on 3-D Digital Imaging and Modeling, 3DIM}.

\bibitem[\protect\citeauthoryear{}{Sacks
  et~al.\@}{2020}]{sacks2020construction}
Sacks, R., Brilakis, I., Pikas, E., Xie, H.~S., and Girolami, M. (2020).
\newblock ``Construction with digital twin information systems.''\ {\em
  Data-Centric Engineering}, 1.

\bibitem[\protect\citeauthoryear{}{Salvi et~al.\@}{2007}]{Salvi2007AEvaluation}
Salvi, J., Matabosch, C., Fofi, D., and Forest, J. (2007).
\newblock ``{A review of recent range image registration methods with accuracy
  evaluation}.''\ {\em Image and Vision Computing}.

\bibitem[\protect\citeauthoryear{}{Sandler
  et~al.\@}{2018}]{sandler2018mobilenetv2}
Sandler, M., Howard, A., Zhu, M., Zhmoginov, A., and Chen, L.-C. (2018).
\newblock ``Mobilenetv2: Inverted residuals and linear bottlenecks.''\ {\em
  Proceedings of the IEEE conference on computer vision and pattern
  recognition},  4510--4520.

\bibitem[\protect\citeauthoryear{}{{Sanicola, L., and Erwin
  S.}}{2021}]{Reuters2021}
{Sanicola, L., and Erwin S.} (2021).
\newblock ``Lack of overhauls at u.s. refiners could stall industry recovery.

\bibitem[\protect\citeauthoryear{}{Savva et~al.\@}{2016}]{savva2016shrec16}
Savva, M., Yu, F., Su, H., Aono, M., Chen, B., Cohen-Or, D., Deng, W., Su, H.,
  Bai, S., Bai, X., et~al.\@ (2016).
\newblock ``Shrec16 track: largescale 3d shape retrieval from shapenet
  core55.''\ {\em Proceedings of the eurographics workshop on 3D object
  retrieval}, Vol.~10.

\bibitem[\protect\citeauthoryear{}{Schnabel
  et~al.\@}{2007}]{Schnabel2007EfficientDetection}
Schnabel, R., Wahl, R., and Klein, R. (2007).
\newblock ``{Efficient RANSAC for Point-Cloud Shape Detection}.''\ {\em
  Computer Graphics Forum}, 26(2), 214--226.

\bibitem[\protect\citeauthoryear{}{Schumaker}{1982}]{Schumaker1982FittingData.}
Schumaker, L.~L. (1982).
\newblock ``{Fitting surfaces to scattered data.}.''\ {\em Proc. NASA workshop
  on surface fitting, College Station, TX, May 1982}.

\bibitem[\protect\citeauthoryear{}{Shepard}{1968}]{Shepard1968AData}
Shepard, D. (1968).
\newblock ``{A two-dimensional interpolation function for irregularly-spaced
  data}.''\ {\em Proceedings of the 1968 23rd ACM National Conference, ACM
  1968}.

\bibitem[\protect\citeauthoryear{}{Stroud}{2006}]{Stroud2006BoundaryTechniques}
Stroud, I. (2006).
\newblock {\em {Boundary Representation Modelling Techniques}}.
\newblock Springer, $<$https://www.springer.com/gp/book/9781846283123$>$.

\bibitem[\protect\citeauthoryear{}{Su
  et~al.\@}{2015}]{Su2015Multi-viewRecognition}
Su, H., Maji, S., Kalogerakis, E., and Learned-Miller, E. (2015).
\newblock ``{Multi-view convolutional neural networks for 3D shape
  recognition}.''\ {\em Proceedings of the IEEE International Conference on
  Computer Vision}.

\bibitem[\protect\citeauthoryear{}{Tangelder and
  Veltkamp}{2008}]{tangelder2008survey}
Tangelder, J.~W. and Veltkamp, R.~C. (2008).
\newblock ``A survey of content based 3d shape retrieval methods.''\ {\em
  Multimedia tools and applications}, 39(3), 441--471.

\bibitem[\protect\citeauthoryear{}{Tchapmi
  et~al.\@}{2017}]{tchapmi2017segcloud}
Tchapmi, L., Choy, C., Armeni, I., Gwak, J., and Savarese, S. (2017).
\newblock ``Segcloud: Semantic segmentation of 3d point clouds.''\ {\em 2017
  international conference on 3D vision (3DV)}, IEEE,  537--547.

\bibitem[\protect\citeauthoryear{}{{Tensorflow}}{2022}]{Tensorflow}
{Tensorflow} (2022).
\newblock ``Transfer learning and fine-tuning.

\bibitem[\protect\citeauthoryear{}{Thomas et~al.\@}{2019}]{thomas2019kpconv}
Thomas, H., Qi, C.~R., Deschaud, J.-E., Marcotegui, B., Goulette, F., and
  Guibas, L.~J. (2019).
\newblock ``Kpconv: Flexible and deformable convolution for point clouds.

\bibitem[\protect\citeauthoryear{}{{Traceparts}}{2019}]{Traceparts2019Traceparts}
{Traceparts} (2019).
\newblock ``{Traceparts}, $<$https://www.traceparts.com/en$>$.

\bibitem[\protect\citeauthoryear{}{{US Department of
  Transportation}}{2016}]{DOT2016}
{US Department of Transportation} (2016).
\newblock ``Distribution, transmission \& gathering, lng, and liquid accident
  and incident data. pipeline and hazardous materials safety administration.

\bibitem[\protect\citeauthoryear{}{Valero
  et~al.\@}{2015}]{Valero2015SemanticTechnology}
Valero, E., Ad{\'{a}}n, A., and Bosch{\'{e}}, F. (2015).
\newblock ``{Semantic 3D reconstruction of furnished interiors using laser
  scanning and RFID technology}.''\ {\em Journal of Computing in Civil
  Engineering}, 30(4), 04015053.

\bibitem[\protect\citeauthoryear{}{Valero
  et~al.\@}{2012}]{Valero2012AutomaticScanners}
Valero, E., Ad{\'{a}}n, A., and Cerrada, C. (2012).
\newblock ``{Automatic Method for Building Indoor Boundary Models from Dense
  Point Clouds Collected by Laser Scanners}.''\ {\em Sensors}, 12(12),
  16099--16115.

\bibitem[\protect\citeauthoryear{}{{VEERUM}}{2022}]{Veerum}
{VEERUM} (2022).
\newblock ``Veerum: Operations, maintenance and reliability.

\bibitem[\protect\citeauthoryear{}{Wang et~al.\@}{2019a}]{wang2019graph}
Wang, L., Huang, Y., Hou, Y., Zhang, S., and Shan, J. (2019a).
\newblock ``Graph attention convolution for point cloud semantic
  segmentation.''\ {\em Proceedings of the IEEE/CVF Conference on Computer
  Vision and Pattern Recognition},  10296--10305.

\bibitem[\protect\citeauthoryear{}{Wang
  et~al.\@}{2018}]{Wang2018SGPN:Segmentation}
Wang, W., Yu, R., Huang, Q., and Neumann, U. (2018).
\newblock ``{SGPN: Similarity Group Proposal Network for 3D Point Cloud
  Instance Segmentation}.''\ {\em Computer Vision and Pattern Recognition}.

\bibitem[\protect\citeauthoryear{}{Wang
  et~al.\@}{2019b}]{Wang2019AssociativelyClouds}
Wang, X., Shen, X., Shen, C., and Jia, J. (2019b).
\newblock ``{Associatively Segmenting Instances and Semantics in Point
  Clouds}.''\ {\em CVPR}.

\bibitem[\protect\citeauthoryear{}{Weinberger and
  Saul}{2009}]{weinberger2009distance}
Weinberger, K.~Q. and Saul, L.~K. (2009).
\newblock ``Distance metric learning for large margin nearest neighbor
  classification..''\ {\em Journal of machine learning research}, 10(2).

\bibitem[\protect\citeauthoryear{}{{White House}}{2021}]{WhiteHouse2022}
{White House} (2021).
\newblock ``Fact sheet: The bipartisan infrastructure deal.

\bibitem[\protect\citeauthoryear{}{{WillowTwin}}{2022}]{Willow}
{WillowTwin} (2022).
\newblock ``Willow: Unlock the power of your building and infrastructure data.

\bibitem[\protect\citeauthoryear{}{Wu
  et~al.\@}{2018}]{Wu2018ConstructingClouds}
Wu, Q., Xu, K., and Wang, J. (2018).
\newblock ``{Constructing 3D CSG Models from 3D Raw Point Clouds}.''\ {\em
  Computer Graphics Forum}.

\bibitem[\protect\citeauthoryear{}{Wu et~al.\@}{2015}]{wu20153d}
Wu, Z., Song, S., Khosla, A., Yu, F., Zhang, L., Tang, X., and Xiao, J. (2015).
\newblock ``3d shapenets: A deep representation for volumetric shapes.''\ {\em
  Proceedings of the IEEE conference on computer vision and pattern
  recognition},  1912--1920.

\bibitem[\protect\citeauthoryear{}{Ye et~al.\@}{2021}]{ye2021learning}
Ye, S., Chen, D., Han, S., and Liao, J. (2021).
\newblock ``Learning with noisy labels for robust point cloud segmentation.''\
  {\em Proceedings of the IEEE/CVF International Conference on Computer
  Vision},  6443--6452.

\bibitem[\protect\citeauthoryear{}{Yin et~al.\@}{2021}]{yin2021automated}
Yin, C., Wang, B., Gan, V.~J., Wang, M., and Cheng, J.~C. (2021).
\newblock ``Automated semantic segmentation of industrial point clouds using
  respointnet++.''\ {\em Automation in Construction}, 130, 103874.

\bibitem[\protect\citeauthoryear{}{Zhao et~al.\@}{2019}]{zhao2019pointweb}
Zhao, H., Jiang, L., Fu, C.-W., and Jia, J. (2019).
\newblock ``Pointweb: Enhancing local neighborhood features for point cloud
  processing.''\ {\em Proceedings of the IEEE/CVF Conference on Computer Vision
  and Pattern Recognition},  5565--5573.

\bibitem[\protect\citeauthoryear{}{Zin{\ss}er
  et~al.\@}{2003}]{Ziner2003AEstimation}
Zin{\ss}er, T., Schmidt, J., and Niemann, H. (2003).
\newblock ``{A refined ICP algorithm for robust 3-D correspondence
  estimation}.''\ {\em IEEE International Conference on Image Processing}.

\end{thebibliography}

%
%
%

\end{document}